\documentclass{article}

    \usepackage[main, final]{neurips_2026}

\usepackage[utf8]{inputenc} 
\usepackage[T1]{fontenc}

\usepackage{amsmath}
\usepackage{amsfonts}

\usepackage[table]{xcolor}
\usepackage{graphicx}
\usepackage{booktabs}
\usepackage{array}
\usepackage{multirow}
\usepackage{threeparttable}
\usepackage{colortbl}

\usepackage{nicefrac}
\usepackage{microtype}
\usepackage{url}
\usepackage{hyperref}

\usepackage{enumitem}
\usepackage{wrapfig}
\usepackage{pifont}

\usepackage{algorithm}
\usepackage{algorithmic}
\newcommand{\algcomm}[1]{\hfill\textcolor{gray}{\scriptsize\textit{// #1}}}
\usepackage{subcaption}

\title{ATI-VLA: Action-Centric Predictive Vision–Language–Action Models via Actionable Alignment Then Adaptive Injection}

\author{%
   \textbf{Yijie Zhu}$^{1,2}$,\,
  \textbf{Rui Shao}$^{1,5}$\thanks{Corresponding authors.},\,
  \textbf{Jie He}$^{1}$,\,
  \textbf{Wei Li}$^{1}$,\,
  \textbf{Bo Zhao}$^{2}$,\,
  \textbf{Yelin Wang}$^{2}$, \\
  \textbf{Xiaochen Yuan}$^{3}$,\,
  \textbf{Tao Tan}$^{3}$,\,
  \textbf{Miao Zhang}$^{1}$,\,
  \textbf{Xiaojiang Peng}$^{4}$,\,
  \textbf{Zitong Yu}$^{2,6}$%
  \footnotemark[\value{footnote}] \\[0.5em]
  $^{1}$Harbin Institute of Technology, Shenzhen \\
  $^{2}$Institute for Artificial Intelligence, Great Bay University \quad
  $^{3}$Macao Polytechnic University \\
  $^{4}$Shenzhen Technology University \quad
  $^{5}$Shenzhen Loop Area Institute \\
  $^{6}$Dongguan Key Laboratory for Intelligence and Information Technology
  \\
  \texttt{zyj99\_hitsz@stu.hit.edu.cn} \quad
  \texttt{shaorui@hit.edu.cn} \quad
  \texttt{yuzitong@gbu.edu.cn} \\
   \texttt{\url{https://jiutian-vl.github.io/ATI-VLA-page}}
}

\begin{document}

\maketitle

\begin{abstract}
  Predictive Vision–Language–Action (VLA) models aim to improve robotic manipulation via future observation or world dynamics forecasting. However, existing approaches often fail to realize this potential and underperform direct action prediction models. We argue that these limitations stem from modality misalignment between observations and actions, together with joint optimization conflicts that drive learning away from an action-centric objective. To this end, we introduce \textbf{\textit{ATI-VLA}}, an Action-Centric Predictive Vision–Language–Action framework via Actionable \textbf{\textit{Alignment}} Then Adaptive \textbf{\textit{Injection}}. Specifically, it follows a two-step design: \textbf{1) \textit{Actionable Representation Alignment via a Shared Codebook}.}
It aligns predictive observation and action representations by mapping both modalities into a shared discrete latent space via a unified codebook, making predictive observation latents readily usable for action generation and mitigating modality misalignment.
 \textbf{2)} \textbf{\textit{Action-Centric Adaptive Injection of Predictive Latents}}. Building upon this, it then injects predictive observation latents into action decoding as explicit predictive priors via a \textbf{\textit{lightweight adaptive side-path}}, enabling adaptive predictive guidance under a single action-centric objective. Extensive experiments on both simulation and real-world robotic tasks demonstrate that ATI-VLA achieves state-of-the-art performance with faster convergence.
\end{abstract}

\section{Introduction}
\begin{figure*}[!h]
  \centering
    \includegraphics[width=1.0\textwidth]{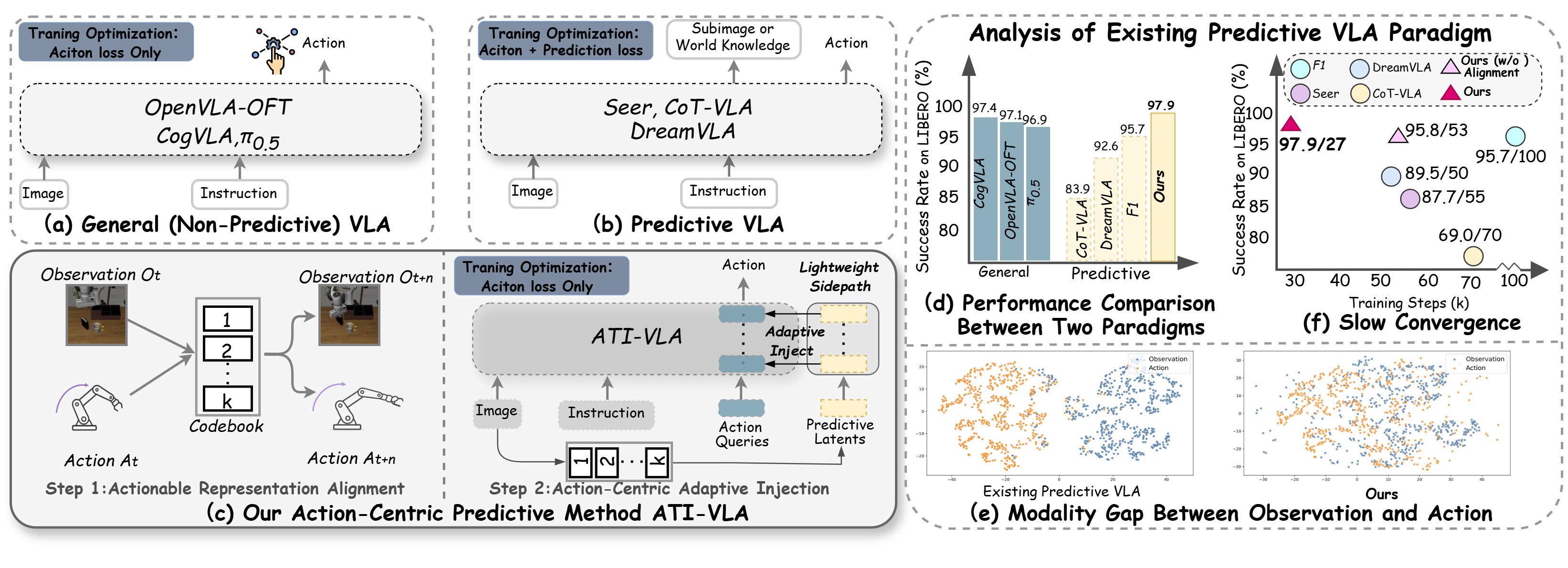}
\caption{
Overview of VLA paradigms and motivation for  \textbf{\textit{ATI-VLA}}.
\textbf{\textit{Figures~(a)--(c)}} compare general (non-predictive) VLA, predictive VLA, and our proposed action-centric predictive framework, \textbf{\textit{ATI-VLA}}.
\textbf{\textit{Figures~(d)--(f)}} identify key limitations of existing predictive VLA approaches, including observation--action modality gaps and slow convergence induced by joint optimization, which are responsible for their inferior performance relative to general VLA.
By addressing these limitations, ATI-VLA enables the predictive VLA paradigm to more fully realize its potential, achieving state-of-the-art performance with faster convergence.
}

    \label{fig1}
    \vspace{-9pt}
    
\end{figure*}

Vision–Language–Action (VLA) models have emerged as a unified paradigm for robotic manipulation, traditionally mapping observations and language instructions directly to low-level actions~\cite{2025pi_05,  yang2025efficientvla, li2025cog, Rt-2, hybridvla, wen2025dexvla, wen2025diffusionvla, lin2025onetwovla}, as shown in Fig.~\hyperref[fig1]{1(a)}. More recently, predictive VLA~\cite{li2025unified, zhu2025h, zhao2025cot_vla, zhang2025dreamvla} approaches extend this paradigm by explicitly forecasting future observations or world dynamics to provide anticipatory guidance for action generation, as shown in Fig.~\hyperref[fig1]{1(b)}.

However, our analysis reveals that existing predictive VLA approaches generally underperform non-predictive VLA baselines, as illustrated in Fig.~\hyperref[fig1]{1(d)}. This indicates that introducing future prediction as an auxiliary task does not reliably translate into effective improvements in action generation. We argue that this performance gap arises because these predictive approaches fail to maintain an action-centric learning focus, primarily due to two factors: \textbf{\textit{i) The Modality Gap Between Visual Observation and Action.}} As shown in Fig.~\hyperref[fig1]{1(e)}, latent distributions analysis of recent predictive VLA models~\cite{zhang2025dreamvla, seer} reveals a pronounced modal gap between observation prediction and action generation. This gap weakens the grounding of predictive representations in action generation, thereby diluting the action-centric learning focus.
 \textbf{\textit{ii) Optimization Conflicts Induced by Joint Training.}}
 Existing predictive VLA paradigms typically introduce future prediction as an auxiliary task by directly adding its loss to the action objective. As illustrated in Fig.~\hyperref[fig1]{1(f)}, such joint optimization leads to unstable training dynamics and slow convergence. Moreover, competing gradients from prediction and action objectives can divert optimization away from action-centric objectives and, in extreme cases, result in training collapse~\cite{zhang2025dreamvla}. Together, these factors hinder predictive VLA approaches from fully realizing their performance potential. \textbf{We provide a more in-depth analysis of these two issues in Appendix~\ref{Modality Gaps}.}

To address these challenges, we propose \textbf{\textit{ATI-VLA}}, an action-centric predictive VLA framework via Actionable \textbf{\textit{Alignment}} Then Adaptive \textbf{\textit{Injection}}, explicitly aligning predictive representations with action generation and injecting predictive information under a single action-centric objective. Specifically, as shown in  Fig.~\hyperref[fig1]{1(c)}, ATI-VLA follows a two-step design: 
\textbf{\textit{1) Actionable Representation Alignment via a Shared Codebook.}} To bridge the modal gap between predictive observations and action generation, ATI-VLA introduces an actionable representation alignment via a shared codebook. By mapping both modalities into a unified discrete latent space, predictive representations are directly grounded in the action space, making them readily compatible with the action decoding process rather than being implicitly entangled through joint training as in prior approaches.
\textbf{\textit{2) Action-Centric Adaptive Injection of Predictive Latents.}} Building upon the aligned and actionable representations, ATI-VLA further introduces an action-centric adaptive injection mechanism to integrate predictive information into action generation. To enable predictive guidance to adapt as action generation progresses, we employ a lightweight adaptive side-path that modulates the action decoding process conditioned on predictive latents. Moreover, to avoid predictive information from overly interfering with action decoding, this modulation is applied at regular layer intervals, allowing predictive guidance while preserving fine-grained control.
Crucially, this injection is optimized solely with the action loss, ensuring that predictive information serves purely as guidance without inducing optimization conflicts.

Extensive experimental results demonstrate that, by addressing the aforementioned challenges, ATI-VLA is able to fully unlock the potential of predictive VLA paradigms.
Across both simulation benchmarks and real-world robotic tasks, ATI-VLA consistently achieves state-of-the-art performance, while exhibiting faster convergence compared to existing predictive approaches. To summarize, our main contributions are as follows:
 \begin{itemize}[leftmargin=*]
    \item We propose \textit{\textbf{ATI-VLA}}, an action-centric predictive VLA framework with an \textit{\textbf{Align-Then-Inject}} design, which effectively addresses the lack of action-centric learning in existing predictive paradigms.
   \item We introduce an \textit{\textbf{actionable representation alignment}} via a shared codebook to ground predictive observations directly in the action space.
    \item We design an \textit{\textbf{action-centric adaptive injection}} strategy that integrates predictive information into action decoding via a lightweight side-path under a single action-centric objective.
    \item Extensive experiments validate the effectiveness of ATI-VLA, achieving state-of-the-art performance with faster convergence.

\end{itemize}
\section{Method}
\subsection{Preliminary: Predictive VLA}
\textbf{The Overall Pipeline.}
Predictive VLA~\cite{zhang2025dreamvla, zhao2025cot_vla, lin2025hif, seer} models improve robot execution by using future perceptual or world knowledge prediction to guide action generation.
At each time step $t$, the robot receives a set of heterogeneous inputs,
including a natural language instruction $\mathcal{I}$, a visual observation $\mathbf{o}_t$,
and two types of special query tokens: the $\mathbf<img>$ token for future prediction
and the $\mathbf<act>$ token for action generation.
All inputs are concatenated into a unified token sequence and processed by a multimodal model $\mathcal{H}$ to produce a predicted latent that encodes future visual content or world knowledge:
\begin{equation}
\small
\mathbf{z}_{t+n}^{\text{pred}}
=
\mathcal{H}\big(\mathbf{o}_t, \mathcal{I}\mid \mathbf<img> \big).
\label{eq:pred_latent}
\end{equation}
Then, a decoder $\mathcal{D}_{\text{img}}$ maps this latent back to the target prediction space, producing either the future image $\hat{\mathbf{o}}_{t+n}$  or the future world knowledge $\hat{\mathbf{w}}_{t+n}$ at time step $t+n$:
\begin{equation}
\small
\hat{\mathbf{y}}_{t+n}=\mathcal{D}_{\text{img}}\!\left(\mathbf{z}_{t+n}^{\text{pred}}\right),
\hat{\mathbf{y}}_{t+n}\in\{\hat{\mathbf{o}}_{t+n},\,\hat{\mathbf{w}}_{t+n}\}.
\end{equation}
For action generation, given the predicted latent $\mathbf{z}_{t+n}^{\text{pred}}$, the $\mathbf<act>$ query is mapped by the unified model $\mathcal{H}$ to a latent action embedding that aggregates action-relevant information from the unified token sequence. Finally, an action head $\mathcal{D}_{\text{act}}$ maps the latent feature to an $n$-step action chunk:
\begin{equation}
\small
\mathbf{A}_{t:t+n-1}
=
\mathcal{D}_{\text{act}}
\Big(
\mathcal{H}\big(\mathbf{o}_t, \mathcal{I},\mathbf<img>\mid \mathbf<act>\big)
\Big).
\label{eq:act_chunk}
\end{equation}
During training, predictive VLA models are typically optimized with a joint objective that supervises both action generation and future prediction:
\begin{equation}
\small
\mathcal{L}
=
\mathcal{L}_{\text{act}}
+
\lambda \, \mathcal{L}_{\text{pred}},
\end{equation}
where $\lambda$ balances action and prediction supervision.

\textbf{Problem Definition.}
As discussed above, this design often introduces a modality gap between predictive and action representations, along with competing optimization objectives that weaken action-centric learning. As a result, predictive information may fail to effectively support action generation and can degrade training stability. The central problem we address is how to leverage predictive representations to enhance action generation while preserving action-centric optimization.

\subsection{ATI-VLA}
\subsubsection{OVERVIEW}
In this section, 
we present ATI-VLA, an \textbf{\textit{Alignment Then Injection}} framework that 
couples Actionable Representation Alignment with Action-Centric Adaptive Injection. ATI-VLA is trained in two stages (Algorithm~\ref{alg:ati-vla}): we first optimize \textbf{Actionable Representation Alignment} to learn a unified, actionable latent space via a shared codebook, making predictive observation latents directly usable for action generation and mitigating modality misalignment. 
We then freeze the Alignment module and train \textbf{Action-Centric Adaptive Injection} with a lightweight adaptive side-path to inject predictive latents into action decoding, optimizing only the action objective. We next elaborate on the two steps in turn, including their architectures and training objectives.

\begin{algorithm}[t]
\small
\caption{ATI-VLA: Align-Then-Inject Framework}
\label{alg:ati-vla}
\begin{algorithmic}[1]
\REQUIRE Current observation and action $\mathcal{O}_t,\mathcal{A}_t$; future observation and action $\mathcal{O}_{t+n},\mathcal{A}_{t+n}$ ($n$: chunk size);
instruction $\mathcal{I}$, initial action tokens $\mathbf{T}_{\text{act}}$;
injection frequency $s$, action length $l_a$; Large language model $\mathbf{H}$ ($L$ layers, layer-$l$ hidden states $\mathbf{H}_l$); layer-$l$ injection weight $\beta_l$. 

\STATE \textcolor{gray}{\textbf{\textit{Step 1: Actionable Representation Alignment}}}
\STATE $\mathbf{Z}_{\mathcal{O}_t}\!\leftarrow\!\text{Quant}(\text{Encode}(\mathcal{O}_t)),\;
\mathbf{Z}_{\mathcal{A}_t}\!\leftarrow\!\text{Quant}(\text{Encode}(\mathcal{A}_t))$
\algcomm{Quantize both modalities via a shared codebook}

\STATE $\hat{\mathcal{O}}_{t+n},\hat{\mathcal{A}}_{t+n}
\leftarrow \text{UnifiedDecoder}([\mathcal{O}_t,\mathcal{A}_t],
[\mathbf{Z}_{\mathcal{O}_t},\mathbf{Z}_{\mathcal{A}_t}])$

\STATE $\mathcal{L}_{\text{align}}\!\leftarrow\!
\lambda_{\mathcal{O}}\mathcal{L}_{\text{obs}}(\mathcal{O}_{t+n},\hat{\mathcal{O}}_{t+n})
+\lambda_{\mathcal{A}}\mathcal{L}_{\text{act}}(\mathcal{A}_{t+n},\hat{\mathcal{A}}_{t+n})$; update $\theta_{\text{Align}}$;
freeze trained $\theta_{\text{Align}}$ after Step 1

\STATE \textcolor{gray}{\textbf{\textit{Step 2: Action-Centric Adaptive Injection}}}
 \STATE $\mathbf{Z}_{\mathcal{O}_t}\!\leftarrow\!\text{Align}(\mathcal{O}_t),\;
    \mathbf{H}_0\!\leftarrow\![\mathcal{O}_t,\mathcal{I},\mathbf{T}_{\text{act}}]$
    
    \FOR{$l=1$ \textbf{to} $L$}
        \STATE $\mathbf{H}_l \leftarrow \text{LLMLayer}_{l}(\mathbf{H}_{l-1})$
        \IF{$l \bmod s = 0$}
            \STATE $\mathbf{H}_l^{\text{act}}\!\leftarrow\!\mathbf{H}_l[:,-l_a:,:],\;
            \mathbf{Z}_{\mathcal{O}_t}\!\leftarrow\!\mathbf{Z}_{\mathcal{O}_t}+\text{FFN}(\mathbf{Z}_{\mathcal{O}_t})$ \algcomm{$\mathbf{Z}_{\mathcal{O}_t}$ is transformed via  side-path for injection}
            
            \STATE $\mathbf{H}_l^{\text{cross}}\!\leftarrow\!\text{CrossAttn}(\mathbf{H}_l^{\text{act}},\mathbf{Z}_{\mathcal{O}_t}),\;
            \mathbf{H}_l[:,-l_a:,:]\!\leftarrow\!\mathbf{H}_l^{\text{act}}+\beta_l\mathbf{H}_l^{\text{cross}}$
        \ENDIF
    \ENDFOR
    
    \STATE $\hat{\mathcal{A}}_{t:t+n-1}\!\leftarrow\!\text{ActionHead}(\mathbf{H}_L[:,-l_a:,:]),\;
    \mathcal{L}_{\text{action}}\!\leftarrow\!\mathcal{L}_{\text{act}}(\mathcal{A}_{t:t+n-1},\hat{\mathcal{A}}_{t:t+n-1})$
\end{algorithmic}
\end{algorithm}
\begin{figure*}
  \centering
    \includegraphics[width=1.0\textwidth]{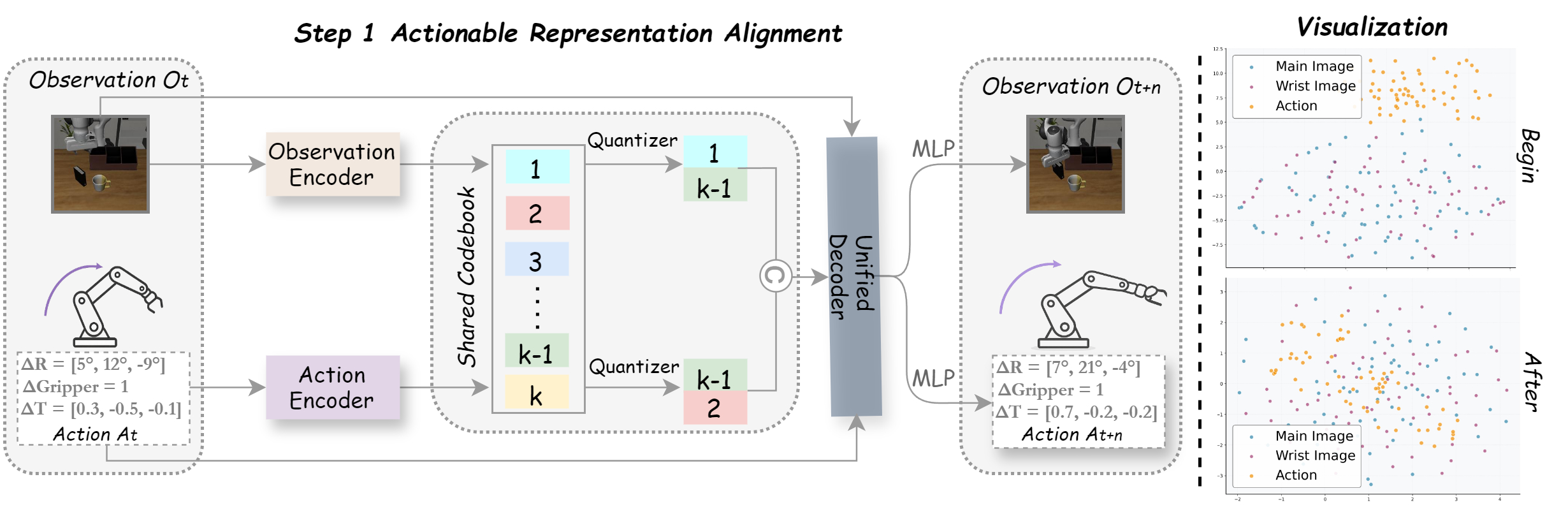}
\caption{\textbf{Step 1: Actionable Representation Alignment.}
\textbf{Left:} The current observation and action are encoded and vector-quantized into a shared codebook, then decoded by a unified attention-based decoder to jointly predict the future observation $\mathcal{O}_{t+n}$ and action $\mathcal{A}_{t+n}$.
\textbf{Right:} t\textsc{-SNE} visualization of observation (main/wrist) and action latents during training: they are initially dispersed and modality-separated, but progressively converge into an aligned shared structure after training.}

    \label{step1}
\end{figure*}
\subsubsection{Actionable Representation Alignment}
As discussed above, 
we first bridge the observation--action modality gap in predictive VLA via an explicit alignment step.
Recent works~\cite{chen2025moto, bu2025learning} have explored extracting action semantics from consecutive frames into compact latent representations, motivating structured cross-modal abstractions.
Inspired by these efforts, we propose Actionable Representation Alignment via a Shared Codebook to bridge this modality gap, as illustrated in Fig.~\ref{step1}.
To ensure effective alignment, we enforce temporal consistency and causal alignment between observation and action representations by adopting a symmetric, predictive design across both branches.
Predictive and action representations are mapped into a shared discrete latent via a unified codebook, grounding predictive latents in the action space.

\textbf{Modality-Specific Encoding.}
Specifically, given the current observation--action pair
$(\mathcal{O}_t,\mathcal{A}_t)$ and the future pair
$(\mathcal{O}_{t+n},\mathcal{A}_{t+n})$, where $n$ denotes the action chunk size,
we encode $\mathcal{O}_t$ and $\mathcal{A}_t$ using modality-specific encoders
$f_{\mathcal{O}}(\cdot)$ and $f_{\mathcal{A}}(\cdot)$.
To explicitly align the two modalities, we introduce a shared codebook
$\mathcal{C}=\{\mathbf{c}_k\}_{k=1}^{K}\subset\mathbb{R}^{d}$,
which defines a unified discrete latent space.
Given a continuous embedding $\mathbf{z}\in\mathbb{R}^{d}$,
the vector quantization operator
$\mathrm{Quant}(\cdot)$ maps it to the nearest codeword in $\mathcal{C}$:
\begin{equation}
\small
\mathrm{Quant}(\mathbf{z})
=
\mathbf{c}_{k^{\ast}},\;
k^{\ast}
=
\arg\min
\|\mathbf{z}-\mathbf{c}_k\|_2^2 .
\end{equation}
Using this shared quantization process, the encoded observation and action
representations are mapped into the same discrete latent space:
\begin{equation}
\small
\mathbf{Z}_{\mathcal{O}_t}
=
\mathrm{Quant}\!\left(f_{\mathcal{O}}(\mathcal{O}_t)\right)
,\;\;
\mathbf{Z}_{\mathcal{A}_t}
=
\mathrm{Quant}\!\left(f_{\mathcal{A}}(\mathcal{A}_t)\right).
\label{eq:shared_quant}
\end{equation}
\textbf{Unified Decoding.}
Given the discrete latent representations
$\mathbf{Z}_{\mathcal{O}_t}$ and $\mathbf{Z}_{\mathcal{A}_t}$ obtained from the shared codebook,
we employ a unified decoder composed of stacked self-attention and cross-attention layers
to jointly model future observations and actions.
We first construct the input token sequence by concatenating the current observation
and action, denoted as $\mathbf{X}_t = [\mathcal{O}_t, \mathcal{A}_t]$.
These tokens are fed into self-attention to capture intra- and cross-modal dependencies:
\begin{equation}
\small
\mathbf{H}_t^{\text{self}}
=
\mathrm{SelfAttn}(\mathbf{X}_t)
=
\mathrm{Softmax}\!\left(
\frac{\mathbf{Q}_t \mathbf{K}_t^{\top}}{\sqrt{d}}
\right)
\mathbf{V}_t,
\end{equation}
where $\mathbf{Q}_t$, $\mathbf{K}_t$, and $\mathbf{V}_t$
are the query, key, and value projections of $\mathbf{X}_t$, respectively.
Next, the self-attended representation $\mathbf{H}_t^{\text{self}}$
is used as the query in a cross-attention block,
which attends to the concatenated discrete latents
$\mathbf{Z}_t = [\mathbf{Z}_{\mathcal{O}_t}, \mathbf{Z}_{\mathcal{A}_t}]$
serving as keys and values:
\begin{equation}
\small
\mathbf{H}_{t+n}
=
\mathrm{CrossAttn}(\mathbf{H}_t^{\text{self}}, \mathbf{Z}_t, \mathbf{Z}_t).
\label{eq:cross_attn}
\end{equation}
The resulting hidden representation $\mathbf{H}_{t+n}$
encodes predictive, action-aware semantics grounded in the shared discrete latent space.
Two lightweight projection heads,
$\mathrm{MLP}_{\mathcal{O}}$ and $\mathrm{MLP}_{\mathcal{A}}$,
are then applied to map $\mathbf{H}_{t+n}$
to the future observation and action predictions, respectively:
\begin{equation} \small \hat{\mathcal{O}}_{t+n} = \mathrm{MLP}_{\mathcal{O}}(\mathbf{H}_{t+n}),\; \hat{\mathcal{A}}_{t+n} = \mathrm{MLP}_{\mathcal{A}}(\mathbf{H}_{t+n}). \end{equation}
\textbf{Training Objective.}
Specifically, the training objective consists of two components: an observation prediction loss and an action prediction loss. The overall alignment objective is defined as \begin{equation} \small \mathcal{L}_{\text{align}} = \lambda_{\mathcal{O}} \mathcal{L}_{\text{obs}} \big( \hat{\mathcal{O}}_{t+n}, \mathcal{O}_{t+n} \big) + \lambda_{\mathcal{A}} \mathcal{L}_{\text{act}} \big( \hat{\mathcal{A}}_{t+n}, \mathcal{A}_{t+n} \big), \label{eq:align_loss} \end{equation}
where $\mathcal{L}_{\text{obs}}$ and $\mathcal{L}_{\text{act}}$ measure errors for observation and action prediction, and $\lambda_{\mathcal{O}},\lambda_{\mathcal{A}}$ balance the two terms. This objective encourages the shared discrete latent space to encode temporally consistent, causally aligned semantics. As shown in Fig.~\ref{step1}, the observation and action latents are initially dispersed and modality-separated, but progressively co-locate and consolidate into a coherent shared structure during Alignment training. This trend suggests that our method effectively reduces cross-modal misalignment and grounds predictive latents in action-relevant semantics. \textbf{Beyond empirical evidence, we also provide theoretical grounding for this design in Appendix~\ref{Shared Discrete Codebook}.}

\subsubsection{Action-Centric Adaptive Injection}
\textbf{Design Rationale.}
With actionable and predictive representations learned in Alignment, 
\begin{wrapfigure}{r}{0.58\textwidth} 
  \centering
   \includegraphics[width=\linewidth]{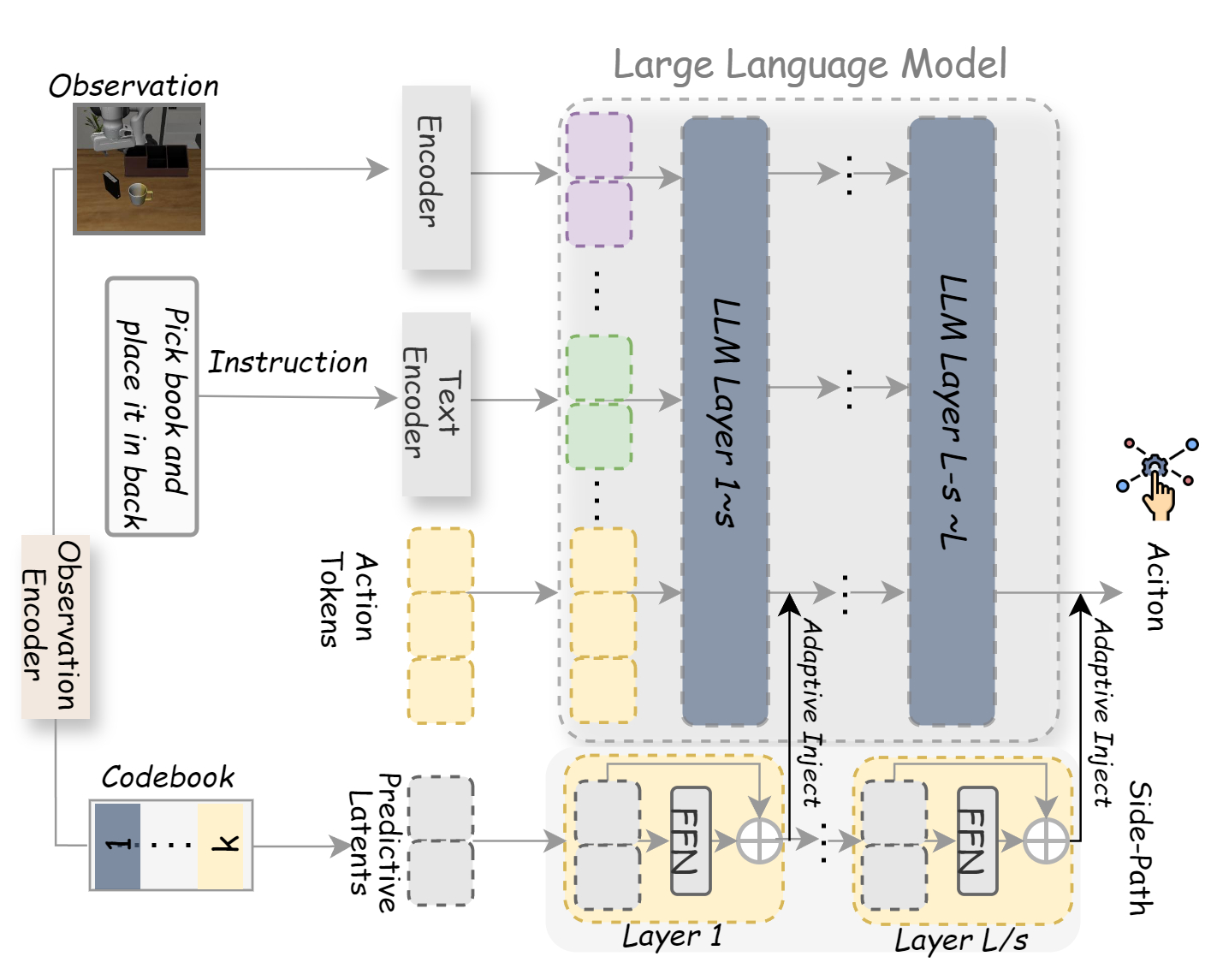}
    \caption{\textbf{Step 2: Action-Centric Adaptive Injection.} Predictive observation latents are adaptively transformed via a lightweight side-path
and injected into LLM layers to modulate action query tokens.
Interval-based injection provides effective predictive guidance
while preserving stable, action-centric decoding.}
\vspace{-5pt}

    \label{injection}
    \end{wrapfigure}
we next integrate predictive information into action generation. 
We use predictive latents purely as guidance while retaining a single action-centric objective. 
A straightforward choice is to inject predictive signals at every decoding layer. 
\textbf{However, our theoretical analysis (Appendix~\ref{Interval Injection}) and empirical results (Tab.~\ref{tab:freq})} show that dense, layer-wise injection can over-constrain decoding, degrading control precision and destabilizing training. This reflects that predictive cues are not uniformly beneficial across the decoding hierarchy: early layers primarily benefit from coarse semantic guidance, whereas later layers require fine-grained motor control. We therefore introduce predictive information \emph{adaptively}, rather than coupling it to every decoding step.

\textbf{Adaptive Side-Path Injection.}
As illustrated in Fig.~\ref{injection},
we introduce a lightweight adaptive side-path that operates in parallel
with a large language model (LLM) backbone,
enabling predictive information to guide action generation
without interfering with the core action-centric optimization.
At each time step $t$, the LLM receives the following inputs:
the current visual observation $\mathcal{O}_t$,
a natural language instruction $\mathcal{I}$,
and a set of learnable action query tokens $\mathbf{T}_{\text{act}}$
with length $l_a$, which are responsible for generating an $n$-step action chunk.
These inputs are first embedded and concatenated to form the initial sequence:
\begin{equation}
\small
\mathbf{H}_0
=
\Big[
f_{\mathcal{O}}(\mathcal{O}_t),\,
f_{\mathcal{I}}(\mathcal{I}),\,
\mathbf{T}_{\text{act}}
\Big],
\end{equation}
where $f_{\mathcal{O}}(\cdot)$ and $f_{\mathcal{I}}(\cdot)$ denote the observation and instruction encoders, respectively. The sequence is then fed into $L$ stacked LLM Layers, yielding hidden states iteratively:
\begin{equation}
\small
\mathbf{H}_l
=
\mathrm{LLMLayer}_l(\mathbf{H}_{l-1}), \qquad l=1,\ldots,L.
\end{equation}
In parallel, we obtain the predictive latent $\mathbf{Z}_{\mathcal{O}_t}$ by encoding $\mathcal{O}_t$ and quantizing it with the shared codebook.
This latent encodes predictive observation information
and serves as the sole source of look-ahead guidance for the LLM. Before each injection, the predictive latent is updated as:
\begin{equation}
\small
\mathbf{Z}_{\mathcal{O}_t}^{(l)}
=
\mathbf{Z}_{\mathcal{O}_t}^{(l-1)}
+
\mathrm{FFN}\!\left(\mathbf{Z}_{\mathcal{O}_t}^{(l-1)}\right),
\end{equation}
with $\mathbf{Z}_{\mathcal{O}_t}^{(0)}=\mathbf{Z}_{\mathcal{O}_t}$.
This transformation adapts the predictive representation to the abstraction level of different LLM layers, instead of keeping it fixed.

Rather than injecting predictive information at every LLM layer,
we perform injection at regular layer intervals to avoid over-constraining
the decoding process.
Concretely, given an injection frequency $s$,
predictive guidance is applied only when $l \bmod s = 0$.
At an injection layer $l$, we extract the hidden states corresponding to
the action query tokens from the LLM output,
denoted as $\mathbf{H}_l^{\text{act}} \in \mathbb{R}^{l_a \times d}$.
The transformed predictive latent $\mathbf{Z}_{\mathcal{O}_t}^{(l)}$
is then injected via cross-attention:
\begin{equation}
\small
\mathbf{H}_l^{\text{cross}}
=
\mathrm{CrossAttn}
\big(
\mathbf{H}_l^{\text{act}},
\mathbf{Z}_{\mathcal{O}_t}^{(l)},
\mathbf{Z}_{\mathcal{O}_t}^{(l)}
\big),
\end{equation}
where the action token representations serve as queries,
and the predictive latent provides keys and values.
The action-related hidden states are subsequently modulated through
a residual update:
\begin{equation}
\small
\mathbf{H}_l^{\text{act}}
\leftarrow
\mathbf{H}_l^{\text{act}}
+
\beta_l \cdot \mathbf{H}_l^{\text{cross}},
\end{equation}
with $\beta_l$ being a learnable scalar that controls the strength
of predictive guidance at layer $l$. After the final LLM layer, the action query representations
$\mathbf{H}_L^{\text{act}}$ are mapped to an $n$-step action chunk.
Crucially, this stage is trained with only the action loss, without any auxiliary prediction objective. 
\textbf{Moreover, the subsequent empirical results show that the injection frequency \(s\) exhibits cross-backbone generalization and introduces little tuning burden.}
\section{Experiment}



 \begin{table*}[t]
    \caption{\textbf{Simulation Results on LIBERO.} Comparison of task success rates and their ranks.}
    \centering
    \footnotesize
    \setlength{\tabcolsep}{3.5pt}
    \begin{sc}
   \begin{tabular}{l|l|ccccc|c}
    \toprule 
    \multirow{2}{*}{Category} & \multirow{2}{*}{Method} & \multicolumn{5}{c|}{\textbf{Success Rate($\uparrow$)}} & \multirow{2}{*}{\textbf{Rank($\downarrow$)}} \\
    \cline{3-7}
    & & Spatial & Object & Goal & Long & Average & \\
    \cmidrule{1-8}
    \multirow{3}{*}{General VLA} 
    & OpenVLA-OFT~{\textit{[RSS'25]}}~\cite{openvla_oft} & 97.6\% & 98.4\% & \textbf{97.9\%} & 94.5\% & 97.1\% & 2 \\
    & $\pi_{0}$~{\textit{[RSS'25]}}~\cite{black2024pi_0} & 96.8\% & 98.8\% & 95.8\% & 85.2\% & 94.2\% & 7 \\
    & PD-VLA~{\textit{[IROS'25]}}\dag~\cite{pdvla} & 95.5\% & 96.7\% & 94.9\% & 91.7\% & 94.7\% & 6 \\
    \cmidrule{1-8}
    \multirow{9}{*}{Predictive VLA} 
    & ATM~{\textit{[RSS'24]}}~\cite{wen2023any} & 68.5\% & 68.0\% & 77.8\% & 39.3\% & 63.4\% & 11 \\
    & Seer~\textit{[ICLR'25]}~\cite{seer} & - & - & - & 87.7\% & - & - \\
    & DreamVLA~\textit{[NeurIPS'25]}~\cite{zhang2025dreamvla} & 97.5\% & 94.0\% & 89.5\% & 89.5\% & 92.6\% & 8 \\
    & CoT-VLA~\textit{[CVPR'25]}~\cite{zhao2025cot_vla} & 87.5\% & 91.6\% & 87.6\% & 69.0\% & 83.9\% & 9 \\
    & $\mathcal{F}_1$~\textit{[arXiv'25]}~\cite{lv2025f1} & 98.2\% & 97.8\% & 95.4\% & 91.3\% & 95.7\% & 3 \\
    & H-GAR~\textit{[AAAI'26]}~\cite{zhu2025h} & - & - & - & 94.0\% & - & - \\
    & UniVLA~\textit{[RSS'25]}~\cite{bu2025learning} & 96.5\% & 96.8\% & 95.6\% & 92.0\% & 95.2\% & 5 \\
    & WorldVLA~{\textit{[arXiv'25]}}~\cite{cen2025worldvla} & 87.6\% & 96.2\% & 83.4\% & 60.0\% & 81.8\% & 10 \\
    & UniVLA~\textit{[ICLR'26]}~\cite{wang2025unified} & 95.4\% & 98.8\% & 93.6\% & 94.0\% & 95.5\% & 4 \\
    \cmidrule{1-8}
    \rowcolor[HTML]{ECDFF2} 
    \multirow{1}{*}{Predictive VLA} & ATI-VLA (Ours) & \textbf{98.6\%} & \textbf{99.4\%} & 97.2\% & \textbf{96.2\%} & \textbf{97.9\%} & \textbf{1} \\
    \bottomrule
\end{tabular}
    \end{sc}
    \label{tab:libero}

\end{table*}
\subsection{Experimental Setup}
\textbf{Implementation Details.}
Training uses 8× NVIDIA H100 (80GB) GPUs, and real-world inference runs on a single RTX 5090 (32GB) GPU (More details in  \textbf{Appendix \ref{Implementation Details}}).

\textbf{Baselines.}
We compare ATI-VLA with representative state-of-the-art methods from two categories:
\textbf{(1) General VLA models}, represented by OpenVLA-OFT~\cite{openvla_oft} and $\pi_0$~\cite{black2024pi_0};
and \textbf{(2) Predictive VLA models}, represented by $\mathcal{F}_1$~\cite{lv2025f1} and DreamVLA~\cite{zhang2025dreamvla}. We further include an ablated variant, \textbf{ATI-VLA w/o alignment}, which jointly models future observations and actions in a single framework, omitting the explicit actionable representation alignment stage. For real-world experiments, we reproduce representative baselines (OpenVLA-OFT~\cite{openvla_oft}, UniVLA~\cite{bu2025learning}, DreamVLA~\cite{zhang2025dreamvla}, and $\mathcal{F}_1$~\cite{lv2025f1}) under the same training and deployment settings as ATI-VLA. More details in \textbf{Appendix \ref{Baselines}}.

\begin{table*}
    \begin{minipage}[t]{0.55\textwidth}
    \centering
    \caption{\textbf{Simulation Results on RoboTwin~2.0 Benchmark.} Comparison of task success rates. ``\dag'' indicates our reproduced results.}
    \scriptsize
    \setlength{\tabcolsep}{0.8pt}
    \begin{sc}
    \begin{tabular}{@{}l|ccccc@{}}
    \toprule 
    \multirow{2}{*}{Method} & \multicolumn{5}{c}{\textbf{Success Rate($\uparrow$)}} \\
    \cline{2-6}
    & Click B & Open L. & Place C.P. & Place B.S. & Avg. \\
    \midrule
    RDT~\cite{liu2024rdt} &80.0\% &59.0\% &78.0\% &5.0\% &56.0\% \\
    ACT~\cite{zhao2023learning} &58.0\% &56.0\% &72.0\% &7.0\% &48.3\% \\
    DP~\cite{chi2025diffusion} &54.0\% &49.0\% &41.0\% &11.0\% &38.8\% \\
    $\pi_0$~\cite{black2024pi_0} &44.0\% &85.0\% &88.0\% &23.0\% &60.0\% \\
    OpenVLA-OFT\dag~\cite{openvla_oft} &78.0\% &66.0\% &82.0\% &9.0\% &58.8\% \\
    \midrule
    \rowcolor[HTML]{ECDFF2} 
    ATI-VLA & \textbf{91.0\%}& \textbf{87.0\%}& \textbf{90.0\%}& 21.0\%& \textbf{72.3\%} \\
    \bottomrule
    \end{tabular}
    \end{sc}
    \label{tab:robottwin}
\end{minipage}
\hfill
\begin{minipage}[t]{0.42\textwidth}
    \centering
\caption{\textbf{Efficiency Comparison Results.}
All methods are evaluated on the same hardware. 
\textit{Cost} denotes training time, with ATI-VLA including both Stage 1 and 2.}
    \scriptsize
    \setlength{\tabcolsep}{2.5pt}
    \begin{tabular}{@{}l|ccc@{}}
    \toprule
    \textbf{Method} & \textbf{Latency}$\downarrow$ & \textbf{Throughput}$\uparrow$  & \textbf{Cost}$\downarrow$  \\
    \midrule
    OpenVLA\dag~\cite{openvla}  & 0.203s & 4.9Hz  & 28.9h  \\
    OpenVLA-OFT\dag~\cite{openvla_oft}& 0.105s & 76.2Hz & 30.2h \\
    DreamVLA\dag~\cite{zhang2025dreamvla} &0.111s  &9.0Hz  &33.5h   \\
    CoT-VLA\dag~\cite{zhao2025cot_vla} &0.130s  &7.7Hz  &67.4h   \\
    \midrule
     \rowcolor[HTML]{ECDFF2} 
    ATI-VLA &0.107s &74.8Hz  &16.2h  \\
     \rowcolor[HTML]{ECDFF2}
    ~~~~\textit{w/o} side-path  & 0.105s & 76.2Hz&- \\
    \bottomrule
    \end{tabular}
    \label{tab:efficiency}
\end{minipage}

\end{table*}
 \begin{table*}[t]
    \caption{\textbf{Real-World Experiments on Different Platforms.}
    The table reports success rates at each stage for five long-horizon tasks, including an out-of-distribution (OOD) variant of T-shirt Folding.
    Results marked with “\dag” were reproduced using the same experimental settings as ATI-VLA. The average success rate corresponds to the completion rate of the entire task sequence.}
    \centering
    \footnotesize
    \setlength{\tabcolsep}{0.5pt}
    \begin{sc}
    \begin{tabular}{l|cc|cc|ccc|ccc|ccc|c}
    \toprule
    \multirow{2}{*}{Method} & \multicolumn{2}{c|}{\textbf{Object P}} & \multicolumn{2}{c|}{\textbf{Markers C}} & \multicolumn{3}{c|}{\textbf{Plate H}}  & \multicolumn{3}{c|}{\textbf{T-shirt F}}& \multicolumn{3}{c|}{\textbf{T-shirt F(OOD)}} & \multirow{2}{*}{Average}\\ 
     & \textbf{Open} & \textbf{+Pack} & \textbf{C1}  & \textbf{+C2}  & \textbf{Pick} & \textbf{+Pass} & \textbf{+Place}  & \textbf{S1} & \textbf{+S2} & \textbf{+S3}& \textbf{S1} & \textbf{+S2} & \textbf{+S3} & \\
    \hline
     \rowcolor[HTML]{EFEFEF} \multicolumn{15}{c}{\textit{\footnotesize
     Galaxea R1 Lite platform}}
     \\
      \hline
    OpenVLA-OFT\dag &18/25&17/25 &17/25&15/25&18/25&17/25&15/25&17/25&15/25&14/25&10/25&8/25&8/25&55\% \\
    UniVLA\dag &14/25&13/25 &13/25&13/25&15/25&14/25&12/25&15/25&15/25&12/25&13/25&10/25&7/25&46\% \\
     $\mathcal{F}_1$\dag &16/25&15/25 &19/25&17/25&15/25&14/25&14/25&17/25&15/25&14/25&16/25&12/25&11/25&57\% \\
     DreamVLA\dag &15/25&14/25 &17/25&14/25&13/25&11/25&11/25&15/25&14/25&14/25&13/25&12/25&10/25&50\% \\
    
        \rowcolor[HTML]{ECDFF2}  
    ATI-VLA(Ours) & \textbf{21/25}&\textbf{19/25} &\textbf{20/25}&\textbf{19/25}&\textbf{20/25}&\textbf{18/25}&\textbf{18/25}&\textbf{19/25}&\textbf{19/25}&\textbf{17/25}&\textbf{17/25}&\textbf{16/25}&\textbf{15/25}&\textbf{70\%} \\
    \hline
       \rowcolor[HTML]{EFEFEF} 
    \multicolumn{15}{c}{\textit{\footnotesize AgileX Cobot Magic platform}} \\
    \hline
        OpenVLA-OFT\dag &16/25&15/25 &19/25&16/25&17/25&17/25&15/25&16/25&15/25&15/25&9/25&8/25&7/25&54\% \\
        $\mathcal{F}_1$\dag &15/25&13/25 &19/25&17/25&15/25&14/25&14/25&17/25&16/25&14/25&15/25&15/25&13/25&57\% \\
            \rowcolor[HTML]{ECDFF2}  
         ATI-VLA(Ours) & \textbf{21/25}&\textbf{20/25} &\textbf{21/25}&\textbf{19/25}&\textbf{19/25}&\textbf{18/25}&\textbf{18/25}&\textbf{20/25}&\textbf{19/25}&\textbf{18/25}&\textbf{18/25}&\textbf{17/25}&\textbf{15/25}&\textbf{72\%} \\
    \bottomrule
    \end{tabular}
    \end{sc}

    \label{tab:real_world}
\end{table*}
\textbf{Simulation \& Real-World Experiments.}
We evaluate ATI-VLA on both simulation benchmarks and real-world robotic tasks.
Simulation experiments are conducted on LIBERO~\cite{libero} (Spatial, Object, Goal, and Long) and RoboTwin 2.0~\cite{chen2025robotwin}.
For real-world evaluation, ATI-VLA is deployed on the Galaxea R1 Lite and AgileX Cobot Magic platforms and tested on five long-horizon tasks across four categories (Object Packing, Markers Collection, Plate Handover, and T-shirt Folding), including an additional out-of-distribution variant for T-shirt Folding.
Each task includes 50 human-teleoperated demonstrations.
Further details in \textbf{Appendix \ref{Simulation Benchmark}} and \textbf{\ref{Real-World Tasks}}.
\subsection{Overall Performance}
\label{sec:overall-performance}

\textbf{Simulation Performance.}
As shown in Tab.~\ref{tab:libero}, ATI-VLA achieves the best overall rank on LIBERO, outperforming both generalist and predictive VLA baselines, with strong results on the Object suite (99.4\%) and Long suite (96.2\%). On RoboTwin~2.0 (Tab.~\ref{tab:robottwin}), ATI-VLA also obtains the highest average success rate, surpassing $\pi_0$ and OpenVLA-OFT, especially on fine-grained manipulation tasks such as Click Bell and Place Container Plate.

\textbf{Real-World Performance.}
On Galaxea R1 Lite, ATI-VLA achieves a \textbf{70\%} average success rate on long-horizon tasks (Tab.~\ref{tab:real_world}), outperforming DreamVLA (50\%), $\mathcal{F}_1$ (57\%), and OpenVLA-OFT (55\%). It also performs well on OOD T-shirt folding (\textbf{60\%}), demonstrating robust real-world generalization and reliable multi-stage control.

\textbf{Efficiency Analysis.} As shown in Tab.~\ref{tab:efficiency}, ATI-VLA achieves the lowest training cost, requiring only 16.2h including both Stage 1 and Stage 2, much less than OpenVLA-OFT (30.2h), DreamVLA (33.5h), and CoT-VLA (67.4h). Its inference efficiency is only marginally lower than OpenVLA-OFT, while the \textit{w/o side-path} variant shows that the side-path adds negligible overhead. 
Meanwhile, ATI-VLA substantially outperforms OpenVLA-OFT with nearly half the training time.

\begin{table*}[t]
\centering

\begin{minipage}[t]{0.5\textwidth}
\centering
\caption{\textbf{Ablation Study on Latent Quality and Alignment.} Top: aligned latents outperform random noise. Bottom: alignment improves performance over unaligned representations.}
\footnotesize
\setlength{\tabcolsep}{1pt}
\begin{sc}
\begin{tabular}{c|cccc}
\toprule
\textbf{Setting} & \textbf{Spatial} & \textbf{Object} & \textbf{Goal} & \textbf{Long} \\
\hline
\rowcolor[HTML]{EFEFEF} \multicolumn{5}{c}{\textit{Injected Latent Quality}} \\
\hline
Random Latent & 95.6\% & 97.0\% & 94.4\% & 88.6\% \\
\rowcolor[HTML]{ECDFF2} 
Aligned Latent & \textbf{98.6\%} & \textbf{99.4\%} & \textbf{97.2\%} & \textbf{96.2\%} \\
\hline
\rowcolor[HTML]{EFEFEF} \multicolumn{5}{c}{\textit{Effect of Representation Alignment}} \\
\hline
ATI-VLA \textit{w/o alignment} & 96.4\% & 97.2\% &95.8\% & 93.6\% \\
\rowcolor[HTML]{ECDFF2} 
ATI-VLA & \textbf{98.6\%} & \textbf{99.4\%} & \textbf{97.2\%} & \textbf{96.2\%} \\
\bottomrule
\end{tabular}
\end{sc}
\label{tab:ablation_latent}
\end{minipage}
\hspace{0.01\textwidth}
\begin{minipage}[t]{0.46\textwidth}
\centering
\caption{\textbf{Ablations on Injection Frequency.}}
\footnotesize
\setlength{\tabcolsep}{2.5pt}
\begin{sc}
\begin{tabular}{c|cccc}
\toprule
\textbf{Frequency} & \textbf{Spatial} & \textbf{Object} & \textbf{Goal} & \textbf{Long} \\
\hline
1 & 95.6\%&97.8\%&94.4\% &92.8\% \\
4 & 98.2\% &98.8\% &96.4\% &93.2\% \\
\rowcolor[HTML]{ECDFF2}
8 &\textbf{98.6\%} & \textbf{99.4\%} & \textbf{97.2\%} & \textbf{96.2\%} \\
16&97.4\% &99.0\% &96.4\% &93.0\% \\
\bottomrule
\end{tabular}
\end{sc}
\label{tab:freq}
\vspace{6pt}
\caption{\textbf{Quantitative Analysis on Actionable Representation
Alignment.} }
\footnotesize
\setlength{\tabcolsep}{1pt}
\begin{sc}
\begin{tabular}{c|cc}
\toprule
\textbf{Setting} & \textbf{Cross-modal Similarity }$\uparrow$ & \textbf{CKNNA}$\uparrow$  \\
\hline
Before & 0.31 & 0.2 \\
\rowcolor[HTML]{ECDFF2}
After & 0.83 & 0.7 \\
\bottomrule
\end{tabular}
\end{sc}
\label{tab:align}
\end{minipage}

\vspace{-15pt}
\end{table*}

\subsection{Ablation Studies}
\label{sec:ablation}
We perform ablation studies on the LIBERO benchmark to validate the effectiveness of our method. 

\begin{wrapfigure}{r}{0.55\textwidth}
  \centering
  \includegraphics[width=0.55\textwidth]{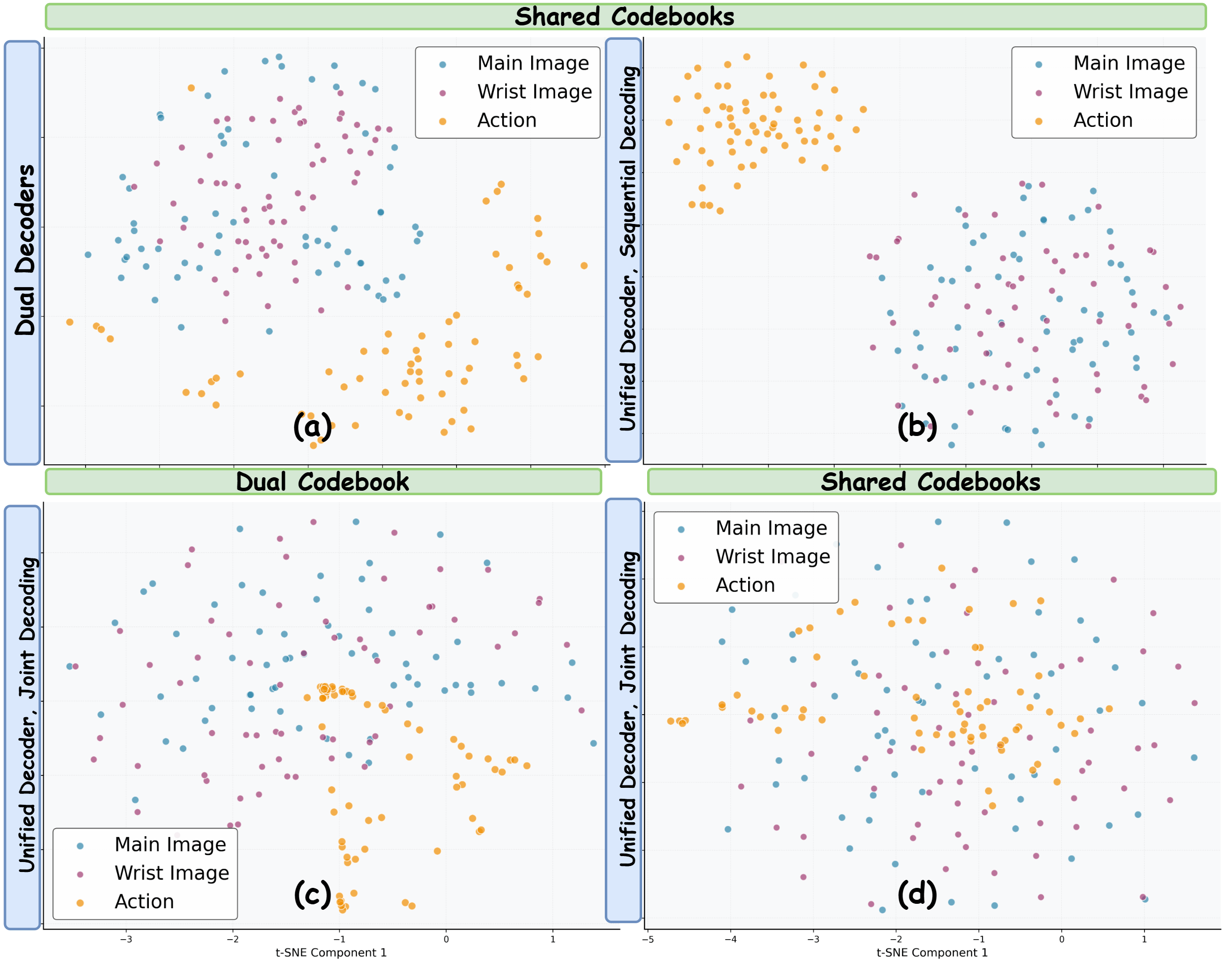}
  \caption{\textbf{Ablation on Actionable Representation Alignment Strategy.} t-SNE of observation/action latents under different alignment strategies: (a) dual decoder, (b) sequential decoding, (c) dual codebook, (d) our method (shared codebook + joint decoding).}
  \label{fig4}
\end{wrapfigure}

\textbf{Ablation on Actionable Representation Alignment Strategy.}
We evaluate the necessity of our shared codebook with joint decoding (Fig.~\hyperref[fig4]{4(d)}) by comparing three variants (Fig.~\hyperref[fig4]{4(a)--(c)}): a) shared codebook with dual decoders, b) shared codebook with sequential decoding, and c) dual codebooks with joint decoding. t-SNE visualizations demonstrate that only our proposed method achieves a well-mixed latent distribution, effectively bridging the observation–action modality gap. 
The quantitative results in Tab.~\ref{tab:align} further confirm this trend, with improved cross-modal similarity and CKNNA after alignment. \textbf{Additional Stage-1 ablations in Appendix Tab.~\ref{tab:stage1_alignment_ablation}} further show the importance of action-side supervision and predictive alignment.

\textbf{Ablation on Latent Representation Quality.}
To assess the role of injected latent quality, 
we compare ATI-VLA with variants that inject \emph{randomly initialized latents} and remove explicit alignment (ATI-VLA \textit{w/o alignment}) in Tab.~\ref{tab:ablation_latent}. 
Injecting random latents substantially degrades performance, indicating that effective guidance requires semantically meaningful signals. Overall, these results suggest that explicit alignment is critical for producing predictive yet action-grounded priors.

\textbf{Ablation on Frequency of Adaptive Injection.}
Tab.~\ref{tab:freq} ablates the injection frequency \(s\). Dense injection (\(s=1\)) hurts long-horizon performance, while overly sparse injection (\(s=16\)) weakens predictive conditioning. The chosen setting \(s=8\) achieves the best trade-off between predictive guidance and action-decoding autonomy. Moreover, Tab.~\ref{tab:backbone_generalization1} shows that the preferred interval remains around one-third to one-fourth of the backbone depth, suggesting cross-backbone generalization with little tuning burden. \textbf{Appendix Tab.~\ref{tab:backbone_generalization} further confirms this trend} across backbones with different architectural priors, including different attention and normalization schemes.

\textbf{Ablation on Isolating the Source of Gains.}
We conduct controlled ablations to isolate the source of performance gains. Tab.~\ref{tab:controlled_same_backbone1} controls for backbone, training data, and evaluation protocol, showing that the improvements mainly come from the full align-then-inject design. \textbf{We further control for different predicted modalities in Appendix Tab.~\ref{tab:predicted_modality}}, confirming that the gains are not due to predicting a particular modality. Together, these results show that the gains mainly come from our align-then-inject design rather than other confounding factors.

\begin{figure*}[t]
    \centering
    \begin{subfigure}[t]{1.0\textwidth}
        \captionsetup{labelformat=empty}
        \includegraphics[width=1.0\textwidth]{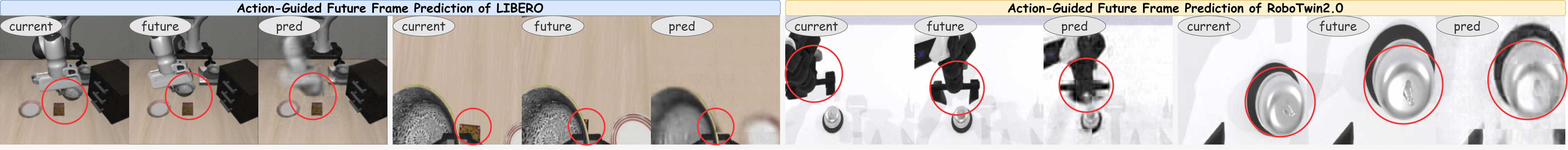}
        \caption{\scriptsize \textbf{(a) Future Observation Prediction from Aligned Action Representations.}
\textbf{Left:} Current observation, ground-truth future, and predicted future (two views) on LIBERO.
\textbf{Right:} RoboTwin 2.0 triplets. Red circles mark dynamic regions linked to imminent state transitions, suggesting the shared latent space captures actionable representations.}

        \label{visualization}
    \end{subfigure}
    \begin{subfigure}[t]{1.0\textwidth}
        \captionsetup{labelformat=empty}
        \includegraphics[width=1.0\textwidth]{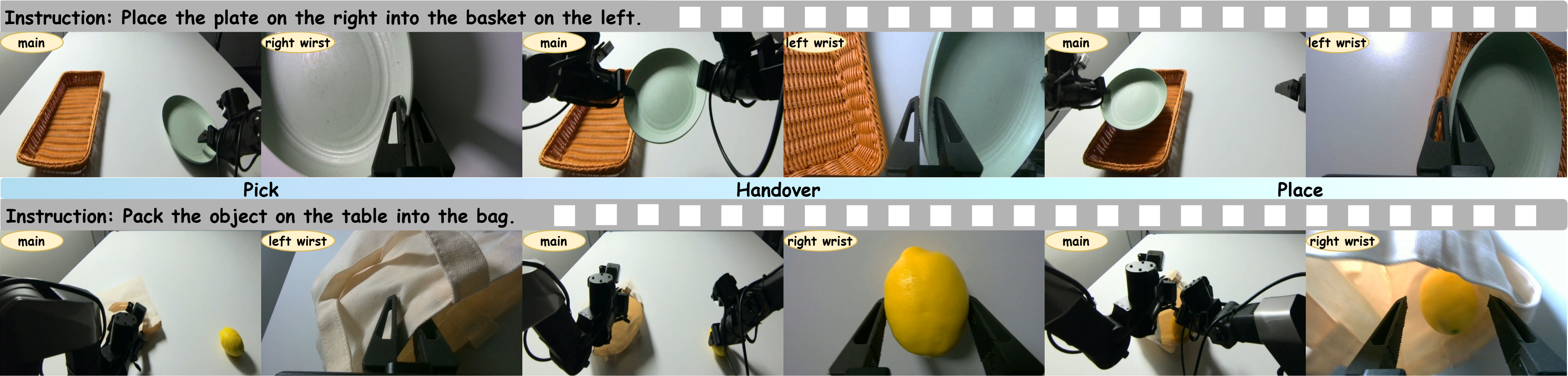}
        \caption{\scriptsize \textbf{(b) Step-by-step Visualization of Two Real-World  Tasks.} \textbf{Additional visualizations are in Appendix~\ref{add_vis}.}}
        \label{vis_real}
    \end{subfigure}
    \caption{Visualization results of the proposed method. (a) shows future observation prediction results. (b) illustrates step-by-step execution of real-world manipulation tasks.}
    \label{fig:combined_results}
    \vspace{-10pt}
\end{figure*}

\begin{table}[t]
\centering
\begin{minipage}[t]{0.48\linewidth}
\centering
\caption{\textbf{Controlled Same-Backbone Ablation.}
All variants share the same setup:
(a) direct future prediction, (b) predictive injection without alignment,
(c) alignment without adaptive injection, and (d) full ATI-VLA.}
\label{tab:controlled_same_backbone1}
\vspace{3pt}
\footnotesize
\setlength{\tabcolsep}{4pt}
\begin{sc}
\begin{tabular}{lc}
\toprule
\textbf{Method} & \textbf{LIBERO} \\
\midrule
Direct Future Prediction & 95.8\% \\
Injection w/o Alignment & 93.4\% \\
Alignment w/o Adaptive Injection & 96.2\% \\
\rowcolor[HTML]{ECDFF2}
ATI-VLA (Ours) & \textbf{97.9\%} \\
\bottomrule
\end{tabular}
\end{sc}
\end{minipage}
\hfill
\begin{minipage}[t]{0.46\linewidth}
\centering
\caption{\textbf{Cross-Backbone Generalization.}
The preferred injection interval $s$ remains near one-third to one-fourth of model depth.}
\label{tab:backbone_generalization1}
\vspace{3pt}
\footnotesize
\setlength{\tabcolsep}{11pt}
\begin{sc}
\begin{tabular}{lcc}
\toprule
\textbf{LLM} & \textbf{Layers $L$} & \textbf{Best $s$} \\
\midrule
LLaMA-2-7B & 32 & 8 $(1/4)$ \\
Qwen2.5-1.5B & 28 & 7 $(1/4)$ \\
Qwen2.5-0.5B & 24 & 8 $(1/3)$ \\
Qwen2.5-3B & 36 & 12 $(1/3)$ \\
\bottomrule
\end{tabular}
\end{sc}
\end{minipage}
\end{table}

\subsection{Qualitative Analysis}
\label{sec:analysis}
The predicted observation frames in Fig.~\ref{visualization}, generated from aligned action representations, exhibit strong structural fidelity and temporal coherence with the ground truth, indicating that the shared latent space captures actionable scene dynamics. The real-world execution sequences in Fig.~\ref{vis_real} further show reliable multi-stage manipulation, such as bag opening, inter-arm plate transfer, and precise placement, demonstrating the practical applicability of ATI-VLA. \textbf{Additional qualitative results are in Appendix~\ref{add_vis}, with real-world videos in the supplementary material.}

\begin{figure*}
  \centering
    \includegraphics[width=1.0\textwidth]{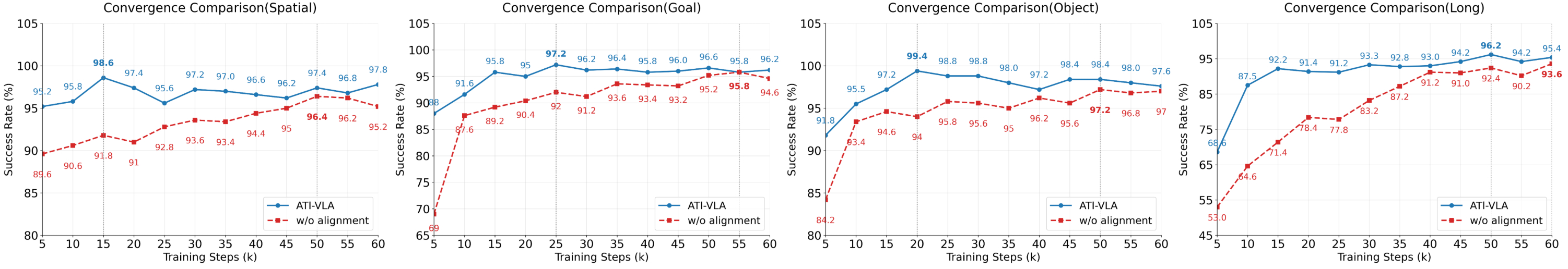}
\caption{\textbf{Convergence Comparison.} ATI-VLA converges faster and performs better than ATI-VLA w/o alignment, while Fig.~\hyperref[fig1]{1(f)} further shows faster convergence over prior predictive VLA baselines.}
    \label{convergence}
\vspace{-10pt}
\end{figure*}

The convergence curves in Fig.~\ref{convergence} show that ATI-VLA converges faster and achieves higher success rates than ATI-VLA w/o alignment, especially on long-horizon tasks. Compared with prior general and predictive VLA baselines, ATI-VLA also exhibits faster optimization and better final performance \textbf{(Fig.~\hyperref[fig1]{1(f)})}. These results suggest that actionable representation alignment provides stable predictive guidance and improves training efficiency.
\section{Related Work}
\textbf{Vision–Language–Action Models.} Vision–Language Models (VLMs)~\cite{flamingo, blip2, xu2025qwen3, llama3, comanici2025gemini,
peng2026survey, xiao2026survey, ye2022unsupervised,shao2024detecting,wang2026rsiccllm, yan2026animeagent, yan2025igdnet,zhu2025emosym,yuan2024auformer, yuan2026multi, zhao2026navgemini, shao2023detecting,shao2019multi} have substantially advanced robotic manipulation by tightly coupling visual perception with linguistic reasoning~\cite{xiao2025ava, li2026global, zhang2026atomicvla,li2026consisvla, shao2025large}. Building on this foundation, RT-2~\cite{Rt-2} pioneered the integration of pre-trained VLMs into Vision–Language–Action modeling, while RT-X~\cite{vuong2023open} further scaled this paradigm with large-scale data to improve generalization across tasks. A series of subsequent approaches, including OpenVLA~\cite{openvla}, OpenVLA-OFT~\cite{openvla_oft}, and the $\pi$-series~\cite{black2024pi_0, 2025pi_05} models, further adopt and extend this paradigm, demonstrating strong performance across a wide range of robotic manipulation tasks.

\textbf{Predictive VLA Models.} 
Recently, increasing attention has been paid to incorporating future prediction into Vision–Language–Action (VLA) models to enhance robotic manipulation~\cite{zhao2025cot_vla, zhang2025dreamvla, lin2025hif, zhang2025up, liu2026last, hu2024video}. Some methods~\cite{du2023learning, black2023zero} employ auxiliary image or video generation models to predict future frames and condition action prediction on the generated goals, introducing additional inference latency. Accordingly, subsequent methods aim to unify action prediction with future modeling, ranging from pixel-level forecasting~\cite{lv2025f1, zhu2025unified} to world knowledge prediction, within a single framework. Despite their strengths, these methods often suffer from modality misalignment and joint optimization conflicts that weaken action-centric learning. In contrast, ATI-VLA explicitly aligns predictive latents with actions and injects them under a single action-centric objective.

\textbf{Robotic Manipulation in Learned Latent Spaces.}
Recent research explores compact, structured latent representations for robotic planning and control~\cite{yang2024vq, mete2407quest, lee2024behavior, li2025star}.
VQ-BeT~\cite{lee2024behavior} and STAR~\cite{li2025star} learn discrete action or skill representations to support transformer-based policy learning, while other approaches learn transferable latent skills from internet-scale human videos~\cite{ye2024latent, chen2025moto, bu2025learning, chen2024igor,nikulin2025latent}.
Both lines of work emphasize learning latent action or skill representations, whereas ATI-VLA targets the observation--action modality gap in predictive VLA.
It also shares the use of discrete codebooks with recent action tokenizers~\cite{wang2025vq, dong2026actioncodec}, but differs in the supervision and role of its codebook.
Specifically, it learns shared, action-grounded prototypes by jointly predicting future observations and actions rather than reconstructing inputs.
The resulting observation latents guide action decoding under a single action objective rather than serve as output action tokens. More closely related, VITA~\cite{ma2026unifying} also learns a shared visual--action codebook.
ATI-VLA nevertheless differs in:
(1) \textit{codebook architecture and learning}, using a unified decoder for joint future observation--action prediction rather than separate autoencoders for modality-specific reconstruction;
(2) \textit{policy architecture}, injecting predictive latents through an adaptive side-path rather than generating shared tokens for dual visual--action decoding;
and (3) \textit{optimization}, freezing the alignment module and using only the action loss in Stage~2, whereas VITA~\cite{ma2026unifying} jointly optimizes observation and action losses during co-training.

\section{Conclusion}
\vspace{-6pt}
We propose \textbf{ATI-VLA}, an action-centric predictive VLA framework that overcomes modality misalignment and optimization conflicts via an Actionable Alignment Then Adaptive Injection design. It first aligns observations and actions in a shared discrete latent space, then adaptively injects predictive latents under a single action objective, achieving state-of-the-art performance with faster convergence. This demonstrates how explicit alignment and adaptive injection can fully realize the potential of predictive VLA for robust robotic manipulation.

\vspace{-6pt}
\section*{Acknowledgment}
This study is supported by National Natural Science Foundation of China (Grant No. 62672144, 62636014, 62576076), Guangdong Basic and Applied Basic Research Foundation (Grant No. 2024A1515010147, 2025B1515120017), Natural Science Foundation of Shenzhen City of China (Grant No. JCYJ20250604145700001), CCF-Tencent Rhino-Bird Open Research Fund, and Guangdong Research Team for Communication and Sensing Integrated with Intelligent Computing (Project No. 2024KCXTD047). The computational resources are supported by SongShan Lake HPC Center (SSL-HPC) in Great Bay University.

{\small
\bibliographystyle{plain}
\bibliography{egbib}
}
\newpage


\appendix

\begin{center}
    \Large \textbf{ATI-VLA: Action-Centric Predictive Vision–Language–Action Models via Actionable Alignment Then Adaptive Injection\\
    \vspace{1em}
    Appendix}
\end{center}

This appendix supplements the main paper with implementation details, additional experimental results, and theoretical analyses. It is organized as follows:
\begin{itemize}[leftmargin=*]
    \item Section~\ref{Implementation Details} details the model architectures and training configurations used in simulation and real-world experiments.
    \item Section~\ref{Benchmarks and Baselines} describes the benchmarks, real-world tasks, evaluation protocols, and baselines.
    \item Section~\ref{Extended Quantitative Analyses} presents extended ablation studies, analyses of observation--action modality gaps, optimization conflicts, and failure modes, as well as additional qualitative results.
    \item Section~\ref{Theoretical Grounding} provides theoretical analyses of the shared  codebook and interval-based latent injection.
    \item Sections~\ref{Limitations and Future Work} and~\ref{Broader Impact and Potential Risks} discuss the limitations of ATI-VLA, future research directions, and broader societal implications, including potential deployment risks.
\end{itemize}
\section{Implementation Details.}
\label{Implementation Details}
\subsection{Model Details}
\label{sec:model_details}

\textbf{Structural Details of Actionable Representation Alignment.}
In the Actionable Representation Alignment stage of ATI‑VLA, we implement three core components:
\begin{itemize}
\item \textbf{Observation Encoder} employs a patch‑embedding layer followed by a spatial Transformer to extract visual tokens, which are then quantized into a shared discrete latent space via a unified codebook. It processes a single input frame and outputs a set of discrete latent codes that capture spatial and semantic information of the scene.
\item \textbf{Action Encoder} uses a lightweight MLP to project the current action vector into a spatially‑aligned embedding grid, which is similarly quantized through the same shared codebook. This ensures that action representations share the same discrete latent structure as visual observations, enabling cross‑modal alignment.
\item \textbf{Unified Decoder} is built with stacked self‑attention and cross‑attention layers, taking the quantized latent representations of both observation and action as context to jointly predict the future observation and future action. The self‑attention captures intra‑modal dependencies, while cross‑attention fuses the aligned observation and action latents for coherent multi‑modal prediction.
\end{itemize}

\textbf{Base Framework and Implementation.}
ATI-VLA adopts OpenVLA-OFT~\cite{openvla_oft} as its implementation framework, with OpenVLA-7B~\cite{openvla} serving as the underlying backbone.
We retain the standard OpenVLA-OFT implementation setup and incorporate the proposed actionable representation alignment and adaptive latent injection mechanisms.

\textbf{Inference Mechanism.}
Unlike video-generation-based policies, ATI-VLA does not require explicit future-observation decoding during inference.
Consistent with the use of predictive latent representations in DreamVLA~\cite{zhang2025dreamvla} and Seer~\cite{seer}, ATI-VLA directly exploits predictive information in latent space to guide action generation.
Specifically, in stage 1, the alignment module is trained with both future-observation and action supervision to learn future-aware, action-grounded representations. This module is subsequently frozen but retained at inference time: the current observation is mapped into the learned latent space, and the resulting predictive latent is adaptively injected into the action decoder as an explicit guidance signal. Therefore, although ATI-VLA does not perform an explicit future-observation rollout at test time, the predictive information learned in Stage 1 continues to directly guide action generation through the retained adaptive side-path.

\subsection{Training Details}
\label{sec:training_details}
\textbf{Simulation Training Setup.}
 For the LIBERO benchmark, the network is adapted using Low-Rank Adaptation (LoRA), configured with a rank of 32 and a scaling parameter \(\alpha = 64\). Training proceeds with a batch size of 64, initiated at a learning rate of \(5 \times 10^{-4}\). The action sequence length is set to \(n = 8\). Validation is performed on benchmark tasks at intervals of every 5,000 steps, with the model checkpoint delivering the highest validation performance being selected for final reporting. Regarding the RoboTwin 2.0 benchmark, we utilize a batch size of 32, an initial learning rate of \(5 \times 10^{-4}\), and an action chunk size of 25.

\textbf{Real-World Training Setup.}
For real-world bimanual manipulation, we follow the standard OpenVLA-OFT setup, using one main-camera view and two wrist-camera views with an action chunk length of \(n = 25\). The model is fine-tuned with LoRA, maintaining a rank of 32 and \(\alpha = 64\). The training regimen spans 80,000 steps with a batch size of 32. The learning rate is initially set to \(5 \times 10^{-4}\) and is subsequently decayed to \(5 \times 10^{-5}\) after 50,000 steps. Starting from the 60,000th step, model checkpoints are evaluated every 5,000 steps. The checkpoint exhibiting the best performance is retained as the final model for deployment.

\section{Benchmarks and Baselines}
\label{Benchmarks and Baselines}
\subsection{Simulation Benchmark}
\label{Simulation Benchmark}
\textbf{LIBERO} is a language-conditioned manipulation benchmark that evaluates complementary facets of robotic capability through multiple task suites. 
\begin{itemize}
\item LIBERO-Spatial contains 10 tasks with identical objects arranged in varied layouts, emphasizing spatial reasoning. 
\item LIBERO-Object includes 10 tasks with fixed layouts but different object instances, testing object-level generalization. 
\item LIBERO-Goal comprises 10 tasks sharing object categories and layouts while varying goals, assessing goal-conditioned behavior diversity. 
\item LIBERO-Long consists of 10 challenging long-horizon tasks across diverse objects and scenes, stressing temporal reasoning and sequential control. 
\end{itemize}
We follow the standard protocol and use 50 expert demonstrations per task from the official release.

\textbf{RoboTwin 2.0} provides a large-scale benchmark for bimanual manipulation, built on an asset library of 731 objects across 147 categories. It defines 50 dual-arm tasks with expert trajectories generated via an MLLM-driven simulation pipeline. To enhance robustness and sim-to-real transfer, it incorporates structured domain randomization across clutter, lighting, backgrounds, table height, and instructions. For evaluation, four tasks are selected: \textbf{Click Bell, Open Laptop, Place Container Plate, and Place Bread Skillet}, each evaluated over 100 trials.

\subsection{Real-World Tasks}
\label{Real-World Tasks}
\textbf{Task Descriptions.}
We deploy ATI-VLA on Galaxea R1 Lite and AgileX Cobot Magic platforms for real-world evaluation across five long-horizon sequential manipulation tasks that span four core capability categories: Object Packing (evaluating dual-arm coordination and container manipulation), Markers Collection (testing fine-grained precision manipulation), Plate Handover (assessing dynamic object transfer and bimanual interaction), and T-shirt Folding (examining deformable object handling in both standard and OOD configurations). These tasks collectively challenge fundamental robotic capabilities including precision manipulation, bimanual coordination, bimanual handover, and dynamic adaptation under uncertainty. Tasks requiring dual-arm coordination (Object Packing, Plate Handover, and both T-shirt Folding variants) evaluate spatial-temporal synchronization and inter-arm transfer, while Markers Collection isolates fine motor control in a single-arm precision scenario. Each task is evaluated through progressive stage success rates to assess intermediate performance, with 50 human-teleoperated demonstrations collected per task to ensure sufficient training data diversity. Details are provided below.

\begin{itemize}
    \item \textbf{Object Packing}\\
    \textit{Instruction:``Pack the object on the table into the bag.''} \\
    This dual-arm sequential manipulation task evaluates coordinated bimanual operation for container handling and object packing, requiring precise temporal coordination and spatial reasoning. The progressive stages include: 1) \textit{``Open''} – securely open the bag and maintain its accessible configuration, and 2) \textit{``+Pack''} – accurately placing tabletop item into the bag while maintaining bag stability.

    \item \textbf{Markers Collection}\\
    \textit{Instruction:``Put the markers into the cup.''} \\
    This task assesses the limits of fine-grained precision manipulation through slender object handling and constrained placement, demanding high precision and delicate handling. The two-stage operation involves: 1) \textit{``C1''} – picking up the first marker and inserting it into the cup with precise orientation control, and 2) \textit{``+C2''} – repeating the process for the second marker while avoiding collisions with the first. 

    \item \textbf{Plate Handover}\\
    \textit{Instruction:``Place the plate on the right into the basket on the left.''} \\
    This dual-arm handover task evaluates dynamic object transfer and inter-arm coordination, focusing on seamless spatial-temporal synchronization during object exchange. The three-stage sequence comprises: 1) \textit{``Pick''} – grasping the target plate securely with the right arm, 2) \textit{``+Pass''} – smoothly transferring the plate to the left arm through a coordinated handover maneuver that maintains object stability throughout the exchange, and 3) \textit{``+Place''} – accurately positioning the plate onto the left-side target plate.

    \item \textbf{T-shirt Folding (In-distribution)}\\
    \textit{Instruction:``Fold the T-shirt.''} \\
    This deformable object manipulation task examines fabric handling and structured folding execution, demanding comprehensive multi-step planning and sustained task coherence across complex operation chains. Following the established three-stage protocol: 1) \textit{``Step 1''} – firmly grasping the T-shirt edge and executing the first controlled folding motion across the garment surface, 2) \textit{``Step 2''} – repeating the folding motion to reinforce the fold, and 3) \textit{``Step 3''} – grasping the right side of the T-shirt and folding it over to the left side. 

    \item \textbf{T-shirt Folding (Out-of-Distribution Variant)}\\
    \textit{Instruction:``Fold the T-shirt.''} \\
    This out-of-distribution variant evaluates generalization capabilities when confronted with systematic variations in garment attributes, while preserving the core folding skill. The task follows the identical three-stage folding procedure, but introduces key variations in the T-shirt properties: 1) the garment exhibits differing visual characteristics including color not seen during training, and 2) the T-shirt dimensions and sizes are altered.
\end{itemize}

\textbf{Data Collection and Augmentation.}
Real-world training data for all tasks is collected via human teleoperation, with 50 expert demonstrations per task. To enhance generalization, moderate data augmentation—varying initial object poses, spatial arrangements, and environmental configurations—is systematically applied to Object Packing, Markers Collection, Plate Handover, and the out-of-distribution T-shirt Folding variant. The in-distribution T-shirt folding task maintains consistent fabric type and initial layout to establish a baseline for deformable object manipulation. Augmentation strategies include randomized object positions, container orientations, and initial arm configurations to simulate realistic execution variability.

\subsection{Baselines}
\label{Baselines}
We compare ATI-VLA against representative state-of-the-art methods from two categories.

\textbf{1) General VLA models.}
These methods directly map the current observation and language instruction to actions, without explicit future prediction:
\begin{itemize}
    \item OpenVLA-OFT~\cite{openvla_oft} is an efficiency-optimized variant of OpenVLA that improves inference speed while maintaining strong general-purpose manipulation performance.
    \item $\pi_0$~\cite{black2024pi_0} follows a dual-system design that combines high-level reasoning with low-level control for long-horizon tasks.
    \item CogVLA~\cite{li2025cog} is a Cognition-Aligned Vision-Language-Action framework that leverages instruction-driven routing and sparsification to improve both efficiency and performance.
    \item PD-VLA~\cite{pdvla} is a performance/efficiency-oriented baseline that accelerates action decoding, serving as a strong non-predictive reference.
\end{itemize}

\textbf{2) Predictive VLA models.}
These methods incorporate look-ahead signals (e.g., future observations, latent dynamics, or world states) to guide action generation:
\begin{itemize}

    \item Seer~\cite{seer} learns predictive inverse dynamics by forecasting future visual states and predicting actions conditioned on the forecast, enabling scalable pretraining and sim-to-real adaptation.
    \item DreamVLA~\cite{zhang2025dreamvla} forecasts compact world knowledge (dynamic/spatial/semantic cues) and uses it for inverse dynamics in a perception--prediction--action loop, with structured attention to disentangle signals and a diffusion-based Transformer for action modeling.
    \item CoT-VLA~\cite{zhao2025cot_vla} adds explicit visual chain-of-thought by autoregressively predicting future frames as intermediate visual goals, then generating short action chunks to reach these goals.
    \item $\mathcal{F}_1$~\cite{lv2025f1}  integrates visual foresight into decision making via a modular Mixture-of-Transformers, predicting goal-conditioned future visuals as planning targets and generating actions through foresight-guided inverse dynamics.
    \item H-GAR~\cite{zhu2025h} introduces a hierarchical interaction framework that refines observation--action trajectories in a goal-driven manner for robotic manipulation.
    \item UniVLA~\cite{bu2025learning} learns cross-embodiment VLA policies by deriving task-centric latent action representations from videos, enabling scalable pretraining on heterogeneous (including human) videos and efficient deployment via latent action decoding.

\end{itemize}
We further include an ablated variant, \textbf{ATI-VLA w/o alignment}, which jointly models future observations and actions in a single framework but removes the explicit Actionable Representation Alignment stage, isolating the effect of alignment from other components of the framework.
\begin{table}[t]
\centering
\caption{\textbf{Trade-off Between Model Complexity and Task Performance.}}
\label{tab:model_complexity}
\footnotesize
\setlength{\tabcolsep}{5pt}
\renewcommand{\arraystretch}{1.2}
\begin{tabular}{lcccccc}
\toprule
\textbf{Method} &
\textbf{Parameters} &
\textbf{FLOPs} &
\textbf{LIBERO} &
\textbf{RoboTwin 2.0} &
\textbf{Real World} &
\textbf{Cost (h)} \\
\midrule
OpenVLA-OFT~(Baseline)\dag&
7.71\,B &
8.45\,T &
96.6 &
58.8 &
55.0 &
30.2 \\
\rowcolor[HTML]{ECDFF2}
\textbf{ATI-VLA} &
\shortstack{\textbf{8.04\,B}\\
\textcolor{red!50!black}{(+4.2\%)}} &
\shortstack{\textbf{8.51\,T}\\
\textcolor{red!50!black}{(+0.7\%)}} &
\shortstack{\textbf{97.9}\\
\textcolor{green!50!black}{(+1.3)}} &
\shortstack{\textbf{72.3}\\
\textcolor{green!50!black}{(+13.5)}} &
\shortstack{\textbf{70.0}\\
\textcolor{green!50!black}{(+15.0)}} &
\shortstack{\textbf{16.2}\\
\textcolor{green!50!black}{(-46.4\%)}} \\
\bottomrule
\end{tabular}
\end{table}
\begin{table}
\centering
\caption{\textbf{Controlled Same-Backbone Ablation.} 
All variants use the same backbone, the same training data, and the same evaluation protocol.}
\label{tab:controlled_same_backbone}
\small
\setlength{\tabcolsep}{6pt}
\begin{tabular}{lccc}
\toprule
\textbf{Method} & \textbf{LIBERO} & \textbf{RoboTwin 2.0} & \textbf{Real-World} \\
\midrule
Baseline + Direct Future Prediction & 95.8\% & 52.4\% & 48\% \\
Baseline + injection without alignment & 93.4\% & 51.3\% & 50\% \\
Baseline + with alignment but without adaptive injection & 96.2\% & 60.7\% & 62\% \\
\rowcolor[HTML]{ECDFF2}
ATI-VLA (Ours) & \textbf{97.9\%} & \textbf{72.3\%} & \textbf{70\%} \\
\bottomrule
\end{tabular}
\end{table}

\section{Extended Analyses}
\label{Extended Quantitative Analyses}
\subsection{Extended Ablations}
\textbf{Trade-off Analysis.}
As shown in Table~\ref{tab:model_complexity}, ATI-VLA consistently improves task performance over OpenVLA-OFT with only 0.7\% additional FLOPs, while reducing total training time by 46.4\% (30.2 to 16.2 hours).
These results demonstrate that ATI-VLA improves policy performance while maintaining comparable computational complexity and substantially reducing training cost.

\textbf{Controlled Same-Backbone Ablation.}
To isolate where the performance gains come from, we conduct a controlled same-backbone ablation. All variants use the same backbone, training data, and evaluation protocol, and differ only in their predictive design.
As shown in Table~\ref{tab:controlled_same_backbone}, direct future prediction yields only limited improvement, while injection without alignment even degrades performance, suggesting that unaligned predictive features may interfere with action decoding. Incorporating alignment improves performance, and the full ATI-VLA achieves the best results across both simulation and real-world settings. These results indicate that the gains primarily stem from the combination of actionable alignment and adaptive injection, rather than from extra supervision, model capacity, or training budget.

\begin{table}[t]
\centering
\caption{\textbf{Controlled Comparison over Predicted Modalities.}
All variants use the same backbone, training data, and evaluation protocol, differing only in the predicted modality and predictive design. Results report average success rates.}
\label{tab:predicted_modality}
\small
\setlength{\tabcolsep}{10pt}
\begin{tabular}{lccc}
\toprule
\textbf{Method} & \textbf{LIBERO} & \textbf{RoboTwin 2.0} & \textbf{Real-World} \\
\midrule
Image Prediction (CoT-VLA-like) & 95.2\% & 52.0\% & 50\% \\
World-Knowledge Prediction (DreamVLA-like) & 94.4\% & 50.3\% & 46\% \\
\rowcolor[HTML]{ECDFF2}
ATI-VLA (Ours) & \textbf{97.9\%} & \textbf{72.3\%} & \textbf{70\%} \\
\bottomrule
\end{tabular}
\end{table}
\begin{table}[t]
\centering
\caption{\textbf{Generalization of Injection Interval Across Backbones.}
We evaluate LLM backbones with different attention patterns and normalization schemes. The preferred injection interval remains around one-third to one-fourth of the model depth.}
\label{tab:backbone_generalization}
\small
\setlength{\tabcolsep}{15pt}
\begin{tabular}{lcccc}
\toprule
\textbf{LLM} & \textbf{Architectural Prior} & \textbf{Norm} & \textbf{Layers} & \textbf{Best $s$} \\
\midrule
LLaMA-2-7B & Dense causal attention & RMSNorm & 32 & 8 $(1/4)$ \\
Mistral-7B & Sliding-window attention & RMSNorm & 32 & 8 $(1/4)$ \\
OPT-1.3B & Dense causal attention & LayerNorm & 24 & 8 $(1/3)$ \\
\bottomrule
\end{tabular}
\end{table}

\textbf{Controlled Comparison over Predicted Modalities.} To account for different predicted modalities for a fair comparison, we reproduce two representative variants under the same framework: an image-prediction variant following the spirit of CoT-VLA~\cite{zhao2025cot_vla}, and a world-knowledge-prediction variant following the spirit of DreamVLA~\cite{zhang2025dreamvla}. As shown in Table~\ref{tab:predicted_modality}, both variants underperform ATI-VLA, especially on RoboTwin 2.0 and real-world tasks. This suggests that the gains do not simply come from predicting a particular modality, but from making predictive representations actionable through alignment and injecting them into action decoding in an adaptive, action-centric manner.

\textbf{Generalization of Injection Interval Across Backbones.}
We conduct ablation experiments on the injection interval using LLM backbones with different attention patterns and normalization schemes. The results in Tab.~\ref{tab:backbone_generalization} show that the best interval remains around one-third to one-fourth of the model depth, indicating that this trend is not specific to specific backbone. Overall, dense injection tends to over-constrain action decoding, while overly sparse injection provides insufficient predictive guidance. Thus, interval-based injection reflects a general trade-off, although the exact optimum may slightly vary across backbone families. Importantly, this does not introduce a substantial tuning burden, since the empirical search space is largely narrowed to a small range around one-third to one-fourth of the backbone depth.

\textbf{Ablation on Stage-1 Alignment Design.}
We further analyze the design of Stage-1 alignment in Table~\ref{tab:stage1_alignment_ablation}. Although Stage 2 injects observation-side latents, action information in Stage 1 is still essential because the goal is to learn an action-grounded shared latent space. Compared with observation-only alignment, using both observations and actions improves performance. This indicates that action-side supervision helps observation latents encode control-relevant factors rather than merely visual appearance.
We also compare predictive alignment with a simpler reconstruction-based objective. Reconstruction-based alignment achieves lower performance across all settings. This suggests that reconstruction alone is insufficient to learn future-oriented actionable semantics. In contrast, our predictive alignment objective encourages the shared latent space to capture action-induced state transitions, leading to stronger and more transferable action guidance.
\begin{table}
\centering
\caption{\textbf{Ablation on Stage-1 Alignment Design.}
We examine whether action information and predictive learning are necessary for learning action-grounded latent representations.}
\label{tab:stage1_alignment_ablation}
\small
\setlength{\tabcolsep}{14pt}
\begin{tabular}{lccc}
\toprule
\textbf{Setting} & \textbf{LIBERO} & \textbf{RoboTwin 2.0} & \textbf{Real-World} \\
\rowcolor[HTML]{EFEFEF}
\hline
\multicolumn{4}{c}{\textit{Effect of Action Information in Stage 1}} \\
\hline
Observation-only Alignment & 95.6\% & 64.3\% & 61\% \\
\rowcolor[HTML]{ECDFF2}
Full Alignment (Ours) & \textbf{97.9\%} & \textbf{72.3\%} & \textbf{70\%} \\
\rowcolor[HTML]{EFEFEF}
\hline
\multicolumn{4}{c}{\textit{Effect of Predictive Alignment Objective}} \\
\hline
Reconstruction-based Alignment & 96.0\% & 62.4\% & 59\% \\
\rowcolor[HTML]{ECDFF2}
Predictive Alignment (Ours) & \textbf{97.9\%} & \textbf{72.3\%} & \textbf{70\%} \\
\bottomrule
\end{tabular}
\end{table}

\textbf{Statistical Significance.}
We report the mean and standard deviation of ATI-VLA across random seeds on LIBERO. ATI-VLA achieves a \(97.9\%\) average success rate with only \(0.1\%\) standard deviation, indicating stable performance across seeds. The small variance on all suites further suggests that the reported gains are not due to random fluctuations.

\subsection{Deeper Analysis of Modality Gaps and Optimization Conflicts.}
\label{Modality Gaps}
\subsubsection{Modality Gaps.}
\textbf{(1) What kind of failures occur.} As shown in Table~\ref{tab:failure_modes_alignment}, the non-aligned variant does not fail diffusely; instead, its failures concentrate on concrete and interpretable cases, including wrong target selection, inaccurate approach poses, mis-oriented placements, and failures to switch actions after subgoal transitions. We categorize these instance-level errors into semantic grounding, pose/control grounding, and transition-phase failures. These failure modes directly reflect the need for strong modality alignment, since the policy must translate predictive semantics into the correct next action chunk. This is particularly critical in Plate Handover and T-shirt Folding, where successful execution requires repeated semantic-to-action remapping across multiple subgoals.
\begin{table}[t]
\centering
\caption{\textbf{Failure-Mode Diagnosis in Long-Horizon Real-World Tasks.}
We categorize failures on Plate Handover and T-shirt Folding, with 100 trials per task and 200 trials in total for each method. The non-aligned variant fails more often in semantic grounding, pose/control grounding, and transition-phase stages.}
\label{tab:failure_modes_alignment}
\small
\setlength{\tabcolsep}{10pt}
\begin{tabular}{lcccc}
\toprule
\textbf{Method} & \textbf{Semantic} & \textbf{Pose/Control} & \textbf{Transition} & \textbf{Success} \\
\midrule
w/o Alignment & 26 & 33 & 49 & 92 \\
\rowcolor[HTML]{ECDFF2}
ATI-VLA (Ours) & \textbf{11} & \textbf{19} & \textbf{41} & \textbf{129} \\
\bottomrule
\end{tabular}
\end{table}

\begin{table}[t]
\centering
\caption{\textbf{First Dominant Failure Stage in Long-Horizon Real-World Tasks.}
We report the stage where the first dominant failure occurs on Plate Handover and T-shirt Folding, with 100 trials per task. The non-aligned variant fails more often in later or transition-heavy stages.}
\label{tab:first_failure_stage}
\small
\setlength{\tabcolsep}{4pt}

\begin{minipage}[t]{0.48\linewidth}
\centering
\vspace{2pt}
\textbf{Plate Handover}\\
\begin{tabular}{lcccc}
\toprule
\textbf{Method} & \textbf{Pick} & \textbf{Pass} & \textbf{Place} & \textbf{Success} \\
\midrule
w/o Alignment & 8 & 18 & 29 & 45 \\
\rowcolor[HTML]{ECDFF2}
ATI-VLA (Ours) & \textbf{5} & \textbf{11} & \textbf{22} & \textbf{62} \\
\bottomrule
\end{tabular}
\end{minipage}
\hfill
\begin{minipage}[t]{0.48\linewidth}
\centering
\vspace{2pt}
\textbf{T-shirt Folding}\\
\begin{tabular}{lcccc}
\toprule
\textbf{Method} & \textbf{S1} & \textbf{S2} & \textbf{S3} & \textbf{Success} \\
\midrule
w/o Alignment & 10 & 15 & 28 & 47 \\
\rowcolor[HTML]{ECDFF2}
ATI-VLA (Ours) & \textbf{4} & \textbf{9} & \textbf{20} & \textbf{67} \\
\bottomrule
\end{tabular}
\end{minipage}

\end{table}

\textbf{(2) When the first dominant failure occurs.}
As shown in Table~\ref{tab:first_failure_stage}, the first dominant failures of the non-aligned variant are concentrated in later or transition-heavy stages, such as \textit{Pass}/\textit{Place} in Plate Handover and the later folding stages in T-shirt Folding. These stages require the policy to re-ground updated predictive semantics into a new control regime after completing previous subgoals. In contrast, ATI-VLA reduces failures across all stages and achieves higher final success rates, suggesting that explicit alignment improves the reliability of semantic-to-action remapping in long-horizon execution.

\textbf{(3) Whether alignment is lower immediately before failure.}
As shown in Table~\ref{tab:prefailure_similarity}, using cross-modal similarity as the alignment metric, the non-aligned variant exhibits a clear drop from safe steps to pre-failure steps, with a gap of 0.22. In contrast, ATI-VLA maintains much higher similarity on both safe and pre-failure steps, and reduces the gap to 0.10. This indicates that failures are closely associated with weakened observation--action alignment, especially in high-risk stages immediately before execution errors.
\begin{table}
\centering
\caption{\textbf{Cross-Modal Similarity Before Failure.}
We compare cross-modal similarity on safe steps and pre-failure steps. The non-aligned variant exhibits a larger alignment drop before failure, while ATI-VLA maintains higher alignment in these high-risk regions.}
\label{tab:prefailure_similarity}
\small
\setlength{\tabcolsep}{15pt}
\begin{tabular}{lccc}
\toprule
\textbf{Method} & \textbf{Safe Steps} $\uparrow$ & \textbf{Pre-Failure Steps} $\uparrow$ & \textbf{Gap} $\downarrow$ \\
\midrule
w/o Alignment & 0.59 & 0.37 & 0.22 \\
\rowcolor[HTML]{ECDFF2}
ATI-VLA (Ours) & \textbf{0.80} & \textbf{0.70} & \textbf{0.10} \\
\bottomrule
\end{tabular}
\end{table}
Overall, these analyses show that modality gaps lead to identifiable failures at key stages, including wrong target selection, incorrect approach or placement poses, and subgoal-switching errors. Such failures are especially frequent in multi-stage transition states, where predictive semantics must be converted into a new action regime. ATI-VLA effectively mitigates these failures by maintaining stronger action-grounded predictive representations.

\subsubsection{Optimization Conflicts.}
We further examine whether joint prediction-action training introduces optimization conflicts. Specifically, we compute the gradient cosine similarity between the prediction loss and the action loss on shared parameters. As shown in Table~\ref{tab:gradient_conflict}, both DreamVLA and our direct-future-prediction variant exhibit negative average cosine similarity and a high ratio of negative gradient pairs, indicating that the two objectives often provide conflicting update directions. Applying PCGrad partially reduces this conflict, improving the average cosine similarity from -0.29 to -0.09 and increasing performance on both RoboTwin 2.0 and real-world tasks. This confirms that gradient interference is a real optimization issue. In contrast, ATI-VLA avoids such conflict by design: after Stage-1 alignment, Stage-2 injection is optimized solely with the action loss, leading to the best performance.
\begin{table}[t]
\centering
\caption{\textbf{Gradient Conflict Between Prediction and Action Objectives.}
We measure the gradient cosine similarity between prediction and action losses on shared parameters. PCGrad partially alleviates the conflict, while ATI-VLA avoids it by design and achieves the best performance.}
\label{tab:gradient_conflict}
\small
\setlength{\tabcolsep}{0.5pt}
\begin{tabular}{lcccc}
\toprule
\textbf{Method} 
& $\textbf{Avg. Cos}(g_{\rm pred}, g_{\rm act}) \uparrow$ 
& \textbf{Neg. Ratio} $\downarrow$ 
& \textbf{RoboTwin 2.0} $\uparrow$ 
& \textbf{Real-World} $\uparrow$ \\
\midrule
DreamVLA 
& -0.23 & 43\% & -- & 50\% \\
Direct Future Prediction 
& -0.29 & 39\% & 52.4\% & 48\% \\
Direct Future Prediction + PCGrad 
& -0.09 & 21\% & 57.6\% & 55\% \\
\rowcolor[HTML]{ECDFF2}
ATI-VLA (Align + Inject) 
& \multicolumn{1}{c}{\textbf{Avoided by design}} 
& \multicolumn{1}{c}{\textbf{Avoided by design}} 
& \textbf{72.3\%} & \textbf{70\%} \\
\bottomrule
\end{tabular}
\end{table}

\subsubsection{Why Is the Modality Gap Severe in Predictive VLA Settings.}
Predictive VLA makes the observation--action modality gap more severe because the predicted representation is often learned under an observation-oriented objective but later used for action generation. Specifically, future-observation prediction encourages visually accurate reconstruction of the next scene. However, many visual details, such as background texture, illumination, or pixel-level appearance, are not directly relevant to control. As a result, the predictive latent may encode rich visual information while failing to emphasize action-relevant factors such as target identity, contact state, object pose, or subgoal transition. When such observation-oriented latents are used to guide the action decoder, the policy must implicitly translate them into executable action chunks, which enlarges the modality gap.

This issue is further amplified by joint prediction-action optimization. In conventional predictive VLA training, the shared representation is optimized by both the prediction loss and the action loss. These two objectives can prefer different features: the prediction loss favors visually faithful future states, whereas the action loss favors compact and control-relevant abstractions. As analyzed in our gradient-conflict study, their gradients can point in conflicting directions, pulling the shared representation away from a purely action-centric solution. Therefore, the modality gap in predictive VLA is not only caused by the inherent difference between observations and actions, but also by the mismatch between the prediction objective and the downstream control objective.

ATI-VLA addresses this issue through the align-then-inject design. Stage 1 uses both observation and action information to learn an action-grounded shared latent space, so that predictive latents encode future-oriented but control-relevant semantics. Stage 2 then injects these aligned latents into action decoding under a single action objective, avoiding further prediction-action optimization conflict. This makes predictive information more directly usable for control rather than leaving the policy to bridge the modality gap implicitly.

\begin{figure*}[t]
    \centering
    \includegraphics[width=1.0\textwidth]{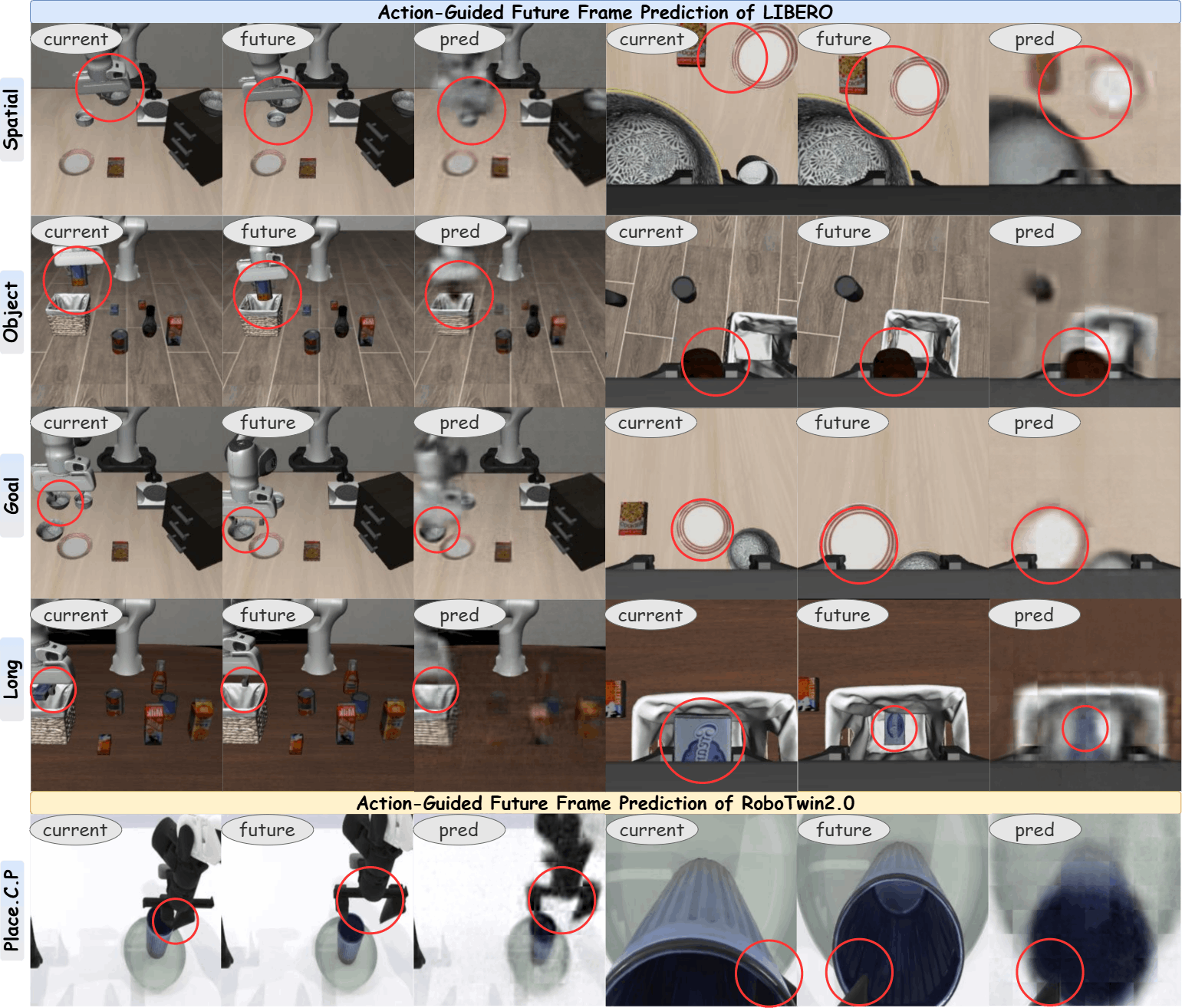}
    \vspace{-17pt}
    \caption{\textbf{Extended Future Observation Prediction from Aligned Action Representations.} Additional qualitative examples of future observation prediction on the LIBERO and RoboTwin 2.0 benchmarks. Each triplet shows the current observation, ground-truth future observation, and the predicted future observation from the aligned action representation. }
    \label{fig:vis_appendix}
    \vspace{-5pt}
\end{figure*}

\begin{figure*}[t]
    \centering
    \includegraphics[width=1.0\textwidth]{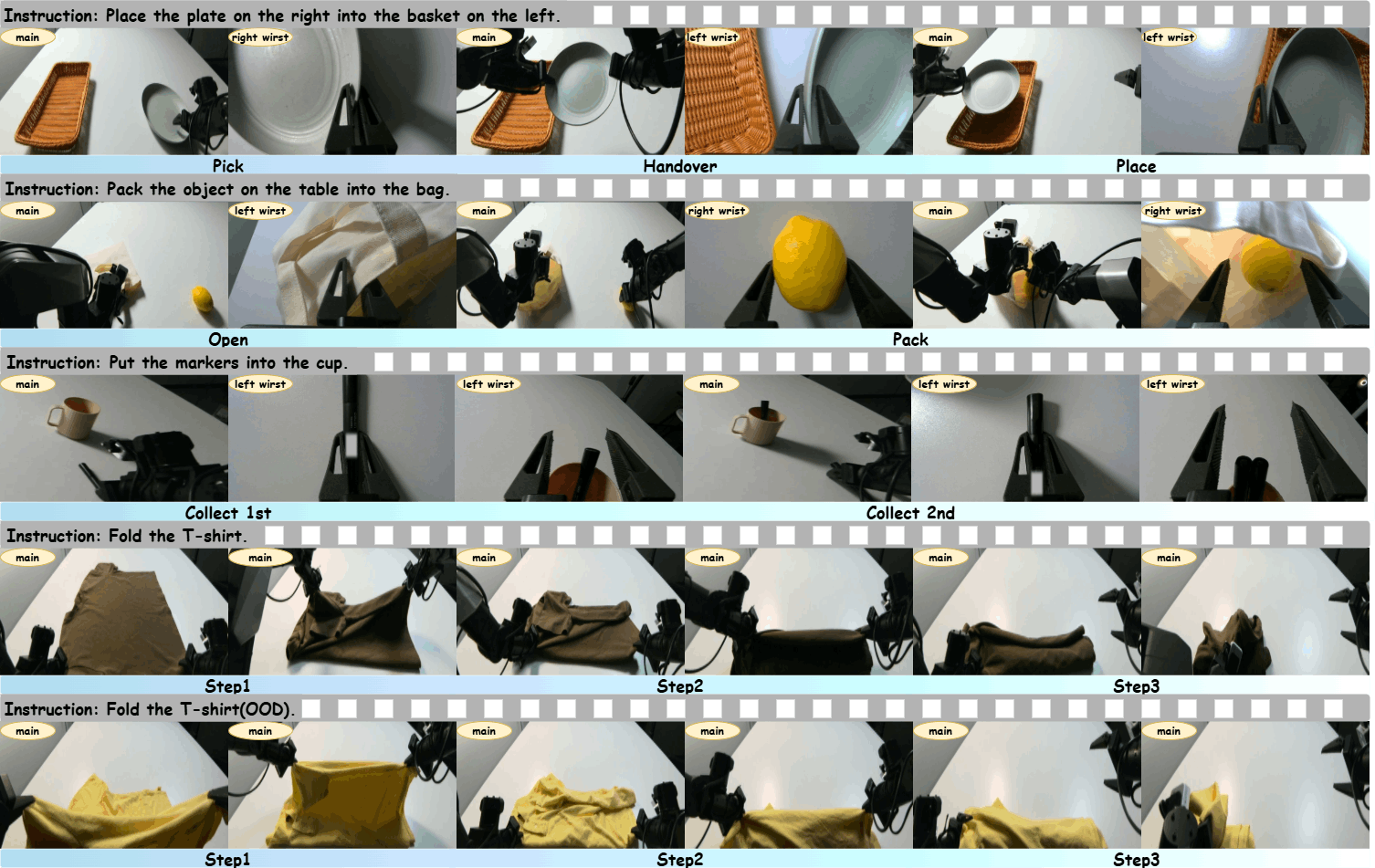}
    \vspace{-17pt}
    \caption{\textbf{Extended Step-by-Step Real-World Task Execution.} Further visualization of all real-world tasks. Each frame captures a critical stage in the sequential execution.}
    \label{fig:vis_real_appendix}
    \vspace{-5pt}
\end{figure*}
\subsection{Additional Qualitative Results}
\label{add_vis}
Figure~\ref{fig:vis_appendix} presents extended qualitative results of future observation prediction generated from aligned action representations on both the LIBERO and RoboTwin 2.0 benchmarks. Across diverse scene layouts, object configurations, and task types, the predicted future frames consistently maintain structural fidelity and temporal coherence with the ground truth. Notably, the predictions preserve fine-grained object details and spatial relationships even in cluttered environments and under varying camera viewpoints. This indicates that the aligned latent space successfully encodes actionable scene dynamics that are both predictive and semantically grounded, effectively bridging the observation-action modality gap. The robustness of these predictions across different task suites further confirms that the alignment strategy yields generalizable representations that transfer well beyond the training distribution.

Figure~\ref{fig:vis_real_appendix} illustrates additional real-world execution sequences of all tasks. In the \textbf{Object Packing} sequence, the model demonstrates reliable dual-arm coordination—first opening the bag with stable bimanual manipulation, then precisely placing the object inside while maintaining bag posture. The \textbf{Markers Collection} sequence highlights fine-grained precision manipulation as the robot picks up slender markers and inserts them into a cup with controlled orientation. The \textbf{Plate Handover} sequence shows smooth inter-arm transfer: securely grasping the plate with the right arm, performing a stable handover to the left arm, and accurately placing it onto the target. The \textbf{T-shirt Folding} sequences (both standard and OOD variants) demonstrate robust deformable object manipulation, maintaining coherent folding motions across varying fabric properties and initial configurations. These sequences collectively highlight the framework's capacity for temporally extended planning, robust bimanual synchronization, and adaptive control under real-world uncertainties.

Together, these extended visualizations reinforce that ATI-VLA's \textit{Align-Then-Inject} design not only produces semantically consistent future predictions but also translates these predictions into stable, precise, and generalizable real-world manipulation behavior. The alignment stage ensures that predictive representations are grounded in the action space, while the adaptive injection mechanism integrates this guidance in a manner that preserves the autonomy and precision of action decoding.

\section{Theoretical Grounding}
\label{Theoretical Grounding}
\subsection{Shared Discrete Codebook}
\label{Shared Discrete Codebook}
Let the observation and action representations be
\[
z_O = f_O(O_t), \qquad z_A = f_A(A_t),
\]
and let the shared codebook be
\[
\mathcal C = \{c_k\}_{k=1}^K, \quad c_k \in \mathbb R^d.
\]
The quantization operator is defined as
\[
Q(z)=c_{k^*(z)}, \quad 
k^*(z)=\arg\min_{k}\|z-c_k\|_2^2,
\]
yielding shared latent representations
\[
\tilde z_O = Q(z_O), \qquad \tilde z_A = Q(z_A).
\]

We focus on the excess action risk induced when predictive observation latents are used for action generation. Let \(\pi(\cdot)\) denote the action decoder and \(\ell(\pi(z),a)\) the loss. Assuming \(\ell(\pi(z),a)\) is \(L\)-Lipschitz with respect to \(z\), we obtain:
\[
\begin{aligned}
\Delta \mathcal R
&:=
\left|
\mathbb E[\ell(\pi(\tilde z_O),a)]
-
\mathbb E[\ell(\pi(\tilde z_A),a)]
\right| \\
&\le
\mathbb E\left[
\left|
\ell(\pi(\tilde z_O),a)-\ell(\pi(\tilde z_A),a)
\right|
\right] \\
&\le
L \cdot \mathbb E\|\tilde z_O-\tilde z_A\|_2.
\end{aligned}
\tag{16}
\]

Equation (16) shows that the excess action risk is upper-bounded by the latent discrepancy between predictive and action representations. Therefore, the role of the shared codebook is to explicitly reduce
\[
\mathbb E\|\tilde z_O-\tilde z_A\|_2,
\]
thereby directly tightening the action risk bound.

Furthermore, the cross-modal discrepancy can be decomposed as
\[
\|z_O-z_A\|_2
\le
\|z_O-\tilde z_O\|_2
+
\|\tilde z_O-\tilde z_A\|_2
+
\|\tilde z_A-z_A\|_2,
\tag{17}
\]
where the middle term
\(\|\tilde z_O-\tilde z_A\|_2\)
represents the discrepancy in a finite prototype space. In particular, if
\[
Q(z_O)=Q(z_A),
\]
then
\[
\tilde z_O=\tilde z_A \;\Rightarrow\; \Delta \mathcal R = 0.
\]

Additionally, the codebook is learned under the joint objective
\[
\mathcal L_{\text{align}}
=
\lambda_O \mathcal L_{\text{obs}}
+
\lambda_A \mathcal L_{\text{act}},
\tag{18}
\]
which enforces both future observation prediction and future action prediction. Hence, the learned prototypes satisfy
\[
\tilde z_O,\tilde z_A \in \mathcal C,
\quad
\min \mathbb E\|\tilde z_O-\tilde z_A\|_2,
\quad
\text{s.t. } \mathcal L_{\text{obs}}+\mathcal L_{\text{act}} \text{ jointly minimized}.
\tag{19}
\]
This explains why the shared discrete codebook reduces the modality gap.
\subsection{Interval Injection}
\label{Interval Injection}
We analyze interval injection as a local first-order perturbation of the action-token decoding trajectory. This local view is important because standard dot-product self-attention is not globally Lipschitz on an unbounded input domain~\cite{kim2021lipschitz}. Instead of assuming global Lipschitzness of the full Transformer, we only require bounded first-order sensitivity in the neighborhood actually traversed by the clean and injected action-token states, which is aligned with recent local analyses of attention and Transformer blocks~\cite{castin2024smooth,havens2024fine,yudin2025pay}.

Let \(\Phi_l\) denote the layer-\(l\) transition restricted to the action-query substate, and let \(\Xi_l\) collect the remaining contextual variables, including observation, language, non-action tokens, normalization statistics, and residual context.
For the clean and injected action-token states \(\widehat H_l^{\mathrm{act}}\) and \(\widetilde H_l^{\mathrm{act}}\), define
\[
E_l := \widetilde H_l^{\mathrm{act}}-\widehat H_l^{\mathrm{act}} .
\tag{20}
\]
We use the following local stability assumption.

\textbf{Assumption (Local Jacobian bound).}
For each layer \(l\), there exists a neighborhood \(\mathcal N_l\) containing the clean and injected action-token states such that
\[
\sup_{U\in\mathcal N_l}
\left\|
\partial_H \Phi_l(U,\Xi_l)
\right\|_2
\le
\rho_l .
\tag{21}
\]
This assumption does not require the full Transformer to be globally Lipschitz. It only bounds the local Jacobian of the action-token transition along the trajectories relevant to ATI-VLA Step 2.

Let the injection layers be
\[
\mathcal I_s := \{l:\ l \bmod s = 0\}.
\tag{22}
\]
At an injection layer \(l\), the predictive side-path contributes
\[
\Delta_l :=
\beta_l\,\mathrm{CrossAttn}\!\left(
\Phi_l(\widetilde H_{l-1}^{\mathrm{act}},\Xi_l),
Z_{\mathcal O_t}^{(l)},
Z_{\mathcal O_t}^{(l)}
\right).
\tag{23}
\]
The clean and injected action-token substates then evolve as
\[
\widehat H_l^{\mathrm{act}}
=
\Phi_l(\widehat H_{l-1}^{\mathrm{act}},\Xi_l),
\qquad
\widetilde H_l^{\mathrm{act}}
=
\Phi_l(\widetilde H_{l-1}^{\mathrm{act}},\Xi_l)
+
\Delta_l\,\mathbf 1_{\{l\in\mathcal I_s\}} .
\tag{24}
\]
By the mean value theorem and the local Jacobian bound,
\[
\|E_l\|_2
\le
\rho_l\|E_{l-1}\|_2
+
\|\Delta_l\|_2\,\mathbf 1_{\{l\in\mathcal I_s\}} .
\tag{25}
\]
Unrolling this recursion yields
\[
\|E_L\|_2
\le
\sum_{l\in\mathcal I_s}
\left(
\prod_{k=l+1}^{L}\rho_k
\right)
\|\Delta_l\|_2 .
\tag{26}
\]

Equation~(26) shows that each injected predictive perturbation is propagated through the remaining action-decoding layers. Since \(|\mathcal I_s|\approx L/s\), dense injection introduces more perturbation sources and can over-condition the action-token trajectory, whereas overly sparse injection provides too few predictive signals to guide decoding effectively. Thus, the injection interval reflects a trade-off between predictive guidance and perturbation accumulation:
\[
\mathcal E_{\mathrm{lack}}(s) \uparrow \text{ with } s,
\qquad
\mathcal E_{\mathrm{perturb}}(s) \downarrow \text{ with } s .
\tag{27}
\]

Therefore, this analysis should be interpreted as a local, trend-level perturbation argument for the adaptive injection stage, rather than as a tight global bound or an exact prediction of the optimal interval. It supports the existence of an intermediate injection regime: denser layer-wise injection accumulates predictive perturbations more strongly through depth, while overly sparse injection weakens predictive guidance. This theoretical trend is consistent with our empirical findings in Table~\ref{tab:backbone_generalization}, where the preferred interval remains around \(L/3\) to \(L/4\) across different LLM backbones.

\section{Limitations and Future Work}
\label{Limitations and Future Work}
Although ATI-VLA shows strong performance on simulation benchmarks and real-world robotic tasks, the current training and evaluation are still based on a relatively limited set of manipulation datasets and task scenarios. In particular, the real-world experiments mainly rely on task-specific teleoperated demonstrations, which validate the effectiveness of our align-then-inject design but do not fully cover the diversity of objects, scenes, embodiments, and interaction patterns in open-world manipulation.
A natural direction for future work is to scale ATI-VLA to larger and more heterogeneous robot datasets. Such data could help the alignment module learn a more comprehensive action-grounded latent space and improve the transferability of predictive latents across tasks and platforms. Future work may also incorporate human videos, cross-embodiment trajectories, and self-supervised interaction data to provide richer priors about object affordances and action-induced state transitions, further extending ATI-VLA toward open-ended long-horizon manipulation.

\section{Broader Impact and Potential Risks}
\label{Broader Impact and Potential Risks}

ATI-VLA aims to improve the reliability and efficiency of VLA models for long-horizon robotic manipulation. By grounding predictive representations in action generation, it may benefit applications such as assistive robotics, warehouse automation, and flexible manufacturing, while also offering a general direction for action-centric predictive learning in embodied AI.

Potential risks may arise when stronger robotic policies are deployed in unconstrained or safety-critical environments, where prediction errors, distribution shifts, or incorrect instruction grounding could lead to unintended behaviors. Practical deployment should therefore include safety constraints, human oversight, permission control, and thorough real-world validation. Future work should further study uncertainty-aware execution and failure detection for safer open-world operation.

\end{document}